\pdfoutput=1
\documentclass[11pt]{article}
\usepackage[]{acl2025_conference}
\usepackage{times}
\usepackage{latexsym}
\usepackage[T1]{fontenc}
\usepackage[utf8]{inputenc}
\usepackage{microtype}
\usepackage{inconsolata}
\usepackage{graphicx}

\usepackage{amsmath,amsfonts,bm}

\def\eqref#1{equation~\ref{#1}}

\def\1{\bm{1}}

\DeclareMathAlphabet{\mathsfit}{\encodingdefault}{\sfdefault}{m}{sl}
\SetMathAlphabet{\mathsfit}{bold}{\encodingdefault}{\sfdefault}{bx}{n}

\usepackage{times}
\usepackage{latexsym}
\usepackage{microtype}      
\usepackage{hyperref}
\usepackage{graphicx}
\usepackage{footmisc}
\usepackage{subcaption}
\usepackage{multirow}
\usepackage{siunitx}
\usepackage{booktabs}
\usepackage{amsthm}
\usepackage{amsmath}
\usepackage{listings}
\usepackage{algorithm}
\usepackage{algorithmic}
\usepackage{lipsum} 
\usepackage{graphicx}
\newcommand{\Comment}[1]{\hfill \textcolor{gray}{\small \(\triangleright\) \textit{#1}}} 
\usepackage{multicol}
\usepackage{CJKutf8}
\usepackage[utf8]{inputenc}
\usepackage{titlesec}
\usepackage[normalem]{ulem}
\usepackage{xspace}
\usepackage{mathtools}
\usepackage{multirow}
\usepackage{pythonhighlight}
\usepackage{wrapfig,lipsum}
\usepackage{inconsolata}
\usepackage{url}            
\usepackage{amsfonts}       
\usepackage{nicefrac}       
\usepackage{tcolorbox}
\usepackage{tabularray}
\usepackage{float}
\usepackage{array}
\usepackage{booktabs}
\usepackage{arydshln}
\usepackage{hhline}
\usepackage[nameinlink]{cleveref}

\newtheorem{hypothesis}{Hypothesis}[section]

\newtheorem{remark}{Remark}[section]
\newtheorem{assum}{Assumption}[section]
\theoremstyle{definition}
\newtheorem{definition}{Definition}[section]
\crefformat{subsection}{\S#2#1#3}
\crefformat{section}{\S#2#1#3}
\crefformat{appendix}{\S#2#1#3}
\newcommand{\triangleq}{\stackrel{\small\triangle}{=}}

\usepackage{caption}
\usepackage{soul}

\newcommand{\llong}[1]{\textcolor{black}{#1}}

\newcommand{\KwStep}[2]{\textbf{\(\bullet\) Step #1:} #2}

\definecolor{inc}{RGB}{84,123,71}
\definecolor{dec}{RGB}{219, 48, 122}
\usepackage[colorinlistoftodos,prependcaption,textsize=tiny]{todonotes}
\usepackage{colortbl}
\newcommand{\model}{LongGuide}

\usepackage{hyperref}
\usepackage{url}

\title{\emph{Beyond In-Context Learning:} Aligning Long-form Generation of Large Language Models via Task-Inherent Attribute Guidelines}

\author{
Do Xuan Long$^{1,3}$, Duong Ngoc Yen$^{2}$, Do Xuan Trong$^{1}$\thanks{Works done during the internship at WING, NUS.}, \\
\textbf{Luu Anh Tuan$^{2}$, Kenji Kawaguchi$^{1}$, Shafiq Joty$^{4}$, Min-Yen Kan$^{1}$, Nancy F. Chen$^{3}$}\\
$^{1}$National University of Singapore,  $^{2}$Nanyang Technological University, Singapore, \\ $^{3}$Institute for Infocomm Research (I$^2$R), A*STAR, $^{4}$Salesforce AI Research \\
\texttt{xuanlong.do@u.nus.edu}
}

\begin{document}

\maketitle

\begin{abstract}
In-context learning (ICL) is an important yet not fully understood ability of large language models (LLMs).  It can greatly enhance task performance using a few examples, termed \emph{demonstrations}, without fine-tuning. Although effective in question answering, ICL often underperforms in long-form generation tasks such as summarization. Under appropriately realistic assumptions, we empirically and theoretically show that ICL demonstrations alone are insufficient to 
teach LLMs the task's language and format distributions for generation. 
We argue for explicit exposure to the task distributions and hypothesize that 
defining them by prompting
enhances model performance. To this end, we present \model{}, 
which efficiently generates two parallel streams of guidelines capturing task language and format properties: 
\emph{(i) Metric Guidelines} (MGs) that instruct models to optimize self-evaluated metrics; and \emph{(ii) Output Constraint Guidelines} (OCGs) that constrain generation at both token and sentence levels. \model{} automatically selects the best combination of guidelines, improving both strong open- and closed-source LLMs by 
over 5\% in both zero- and few-shot settings.
\model{} is generalizable, learnable by weak models to enhance strong ones, and integrates synergistically with automatic prompt optimizers\footnote{Our codes and data will be made available at \href{https://github.com/dxlong2000/LongGuide}{here}.}.

\end{abstract}

\section{Introduction} \label{sec:intro}

In recent years, pre-trained large language models  (LLMs) have demonstrated impressive instruction-based performance through zero- and few-shot learning capabilities \citep{brown2020language,chowdhery2023palm,openai2022chatgpt,touvron2023llama,jiang2023mistral,team2023gemini}. Notably, few-shot learning, termed as in-context learning (ICL), has proven highly effective and widely used to calibrate LLMs for applications \citep{dong2022survey}. Formally, let $\mathcal{V}$ be the vocabulary of the LM. For a test-time task $T$, the goal is to generate a token sequence $y \in \mathcal{Y} \subseteq \mathcal{V}^*$, given input token sequence $x \in \mathcal{X} \subseteq \mathcal{V}^*$. Then, ICL generation using an LLM $\mathcal{M}$ is the generation conditioned on $x$ with $k$ task {demonstrations} $\{(x_1, y_1), ..., (x_k, y_k)\}$ concatenated into $d_f = [x_1, y_1, ..., x_k, y_k] \in \mathcal{D} \subseteq \mathcal{V}^*$.  The probability distribution induced from $\mathcal{M}: \mathcal{V^*} \rightarrow \mathbb{R}$ is:

\vspace{-5mm}
\begin{equation}
\begin{aligned}
\label{eq:gen}
P_{\mathcal{M}}(y | d_f, x) :=\; & \prod_{t=1}^{|y|} \mathcal{M}_{y^t}([x_1, y_1, \dots, \\
& \hspace{15mm} x_k, y_k, x, y^{<t}])
\end{aligned}
\end{equation}

\begin{figure}
\centering
\includegraphics[width=1\columnwidth]{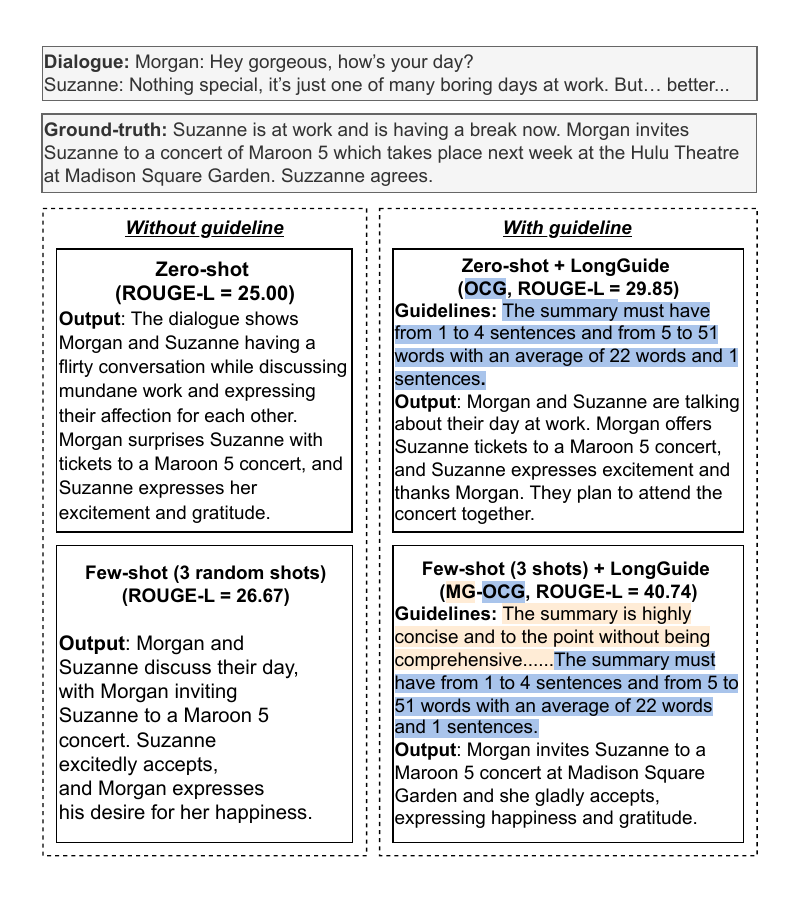}
\caption{\small{ChatGPT results on a SAMSum example \citep{gliwa-etal-2019-samsum}. With \model{}, the generated output aligns better with ground truth, and the quality is also improved by removing verbose details: ``The dialogue...'' (ZS), ``...discuss their day'' (FS). See Appx.-\Cref{fig:teaser-full} for full texts.}}
\label{fig:longguide-teaser}
\end{figure}

where $y = [y^1, \dots, y^{|y|}]$ with $y^t \in \mathcal{V}$.
Several prior studies attempt to explain the ICL capabilities of LLMs, advocating for the sufficiency of well-chosen $d_f$ as implicitly teaching the $\mathcal{M}$ to perform the tasks, especially classification ones \citep{saunshi2020mathematical,xie2021explanation,wang2024large}. Central to their theoretical analyses is a strong assumption that the language model $\mathcal{M}$ fully captures the underlying distribution of the task's language; i.e., $P_{\mathcal{M}}(X) = P_{T}(X)$ where $P_{T}$ is the task-specific data distribution. However, this assumption is often not met, especially with domain-specific terminologies \citep{cheng2024adapting} (also see \Cref{appx:case-study-assum} for a case study), raising concerns about the actual sufficiency of ICL. Moreover, recent studies empirically show that ICL underperforms in long-form generation tasks involving multi-sentence or -paragraph answers \citep{sun-etal-2023-evaluating,huang2024calibrating}, highlighting significant gaps in our understandings of the causes of these limitations and how to effectively instruct LLMs for these tasks. These challenges remain unsolved to date.

In this work, we first study the proficiency of ICL for long-form generation tasks. We empirically and theoretically highlight that if a language model fails to capture the task’s \emph{text properties (language and format)}, providing demonstrations alone with such properties cannot entirely resolve this (\Cref{sec:demos-anayn}). This is because the model does not consistently apply them to all generated responses. Maintaining such properties in responses is crucial for accurately solving the task. Therefore, we argue that providing explicit task guidelines that capture these text properties is essential for improving LLM performance. \Cref{fig:longguide-teaser} illustrates such an example where instructing LLMs explicitly by guidelines carrying certain properties (e.g., conciseness, \#sentences) of the task output distribution improves both alignment with ground truth and generation quality.

We then propose \model{} (\Cref{sec:method}), a guideline-learning algorithm that efficiently\footnote{The prompts and cost analysis are discussed in \Cref{appdx:prompts}.} generates two types of guidelines concurrently from limited task training data as supplementary instructions to enhance LLMs: (i) {Metric Guidelines (MGs)} that steer models to optimize self-evaluation guided metrics, inspired by prior studies in machine translation \citep{ranzato2015sequence} and LLM self-evaluation \citep{ren2023self}; and (ii) {Output Constraint Guidelines (OCGs)} that impose constraints on generated outputs at the sentence and token levels, drawing on controllable generation research \citep{fan-etal-2018-controllable}. \model{} is related to previous studies in task instruction construction \citep{wang-etal-2022-super} and task understanding through task definitions \citep{yin-etal-2023-read}. However, it differs by offering ``post-hoc'' instructions that guide LLMs to enhance responses based on learned quality and quantitative criteria.

 
\model{} automatically identifies optimal guidelines, significantly enhancing distribution alignment and generation quality across seven generation tasks \llong{and one real-life chat LLM benchmark}, including summarization, text simplification, translation, dialogue generation, and table-to-text generation. Its guidelines can enhance ICL performance through demonstrations (\Cref{ssec:generalize-from-demos}), improve non-instruct LLMs (\Cref{sec:can-longguide-improve-non-instruct}), boost stronger models when learned by weaker ones (\Cref{appdx:longguide-is-transferrable}), and can be further optimized for usage using prompt optimization algorithms (\Cref{sec:longguide-can-be-combined}). \llong{Notably, \model{} is approximately at least 3.75 times more cost-efficient than prompt optimization algorithms (\Cref{appx:longguide-prompts}-\Cref{tab:prompt_costs_vs_baselines}) as it requires only four prompt variants to verify on the validation set while delivering superior performance.}

\section{ICL Alone Is Insufficient for Long-form Generation} \label{sec:demos-anayn}


\llong{A long-form generation task $T$ with $n$ samples is defined as $D = \{(x^t_i,y^t_i)\}_{i=1}^{n}$ where $x^t_i$ and $y^t_i$ are input contexts and ground-truth {sentence- or paragraph-long} responses \citep{fan-etal-2019-eli5}}. For such tasks, preserving language and format properties of task-specific data during generation is essential for aligning outputs with ground truth.  This is unlike classification, where outputs are predefined. We demonstrate that ICL fails to enable LLMs to maintain these properties during generation. 

\begin{table*}
\centering
\resizebox{1\textwidth}{!}{ 
\begin{tabular}{l|cccccc|cc} 
\toprule
\textbf{ICL w/ 5 demos} & \textbf{(1) COV} & \textbf{(2) FAC} & \textbf{(3) CON} & \textbf{(4) INF} & \textbf{(5) COH} & \textbf{(6) REL} & \textbf{(7) NT (mean)} & \textbf{(7) NT (std)} \\
\emph{Expected outcome} & \emph{100\%} & \emph{100\%} & \emph{100\%} & \emph{100\%} & \emph{100\%} & \emph{100\%} & \emph{17.00} & \emph{0.00} \\ 
\midrule
Mistral-7B-v0.3 & 12\% & 27\% & 28\% & 8\% & 20\% & 35\% & 87.74 & 144.91 \\ 
Llama-3.1-8B & 12\% & 42\% & 50\% & 4\% & 32\% & 47\% & 271.81 & 379.48 \\ 
Qwen2.5-7B & 43\% & \textbf{90\%} & \textbf{85\%} & \textbf{40\%} & 78\% & \textbf{96\%} & 281.38 & 264.59 \\ 
\midrule
Mistral-7B-it-v0.2 & 38\% & 80\% & 78\% & 17\% & 75\% & 88\% & 50.25 & 55.54 \\ 
Llama-3.1-8B-it & \textbf{44\%} & 86\% & 82\% & 26\% & \textbf{81\%} & 87\% & \textbf{34.72} & \textbf{45.29} \\ 
\bottomrule
\end{tabular}}
\caption{\small{\% of responses scored $5$ on the (1)-(6) metrics, and the (mean, std) of the (7) \#tokens of the responses. \llong{Qwen scored high on metrics (1)--(6) because it copies the input dialogue as the summarization outcome.}}}
\label{tab:icl-motivation-results}
\end{table*}


\paragraph{Setups.} We first select metrics as properties commonly used for dialogue summarization. We follow \citet{fu2023gptscore} to choose six: \textbf{(1) Semantic Coverage (COV)}; \textbf{(2) Factuality (FAC)}; \textbf{(3) Consistency (CON)}; \textbf{(4) Informativeness (INF)}; \textbf{(5) Coherence (COH)}; \textbf{(6) Relevance (REL)}. We also measure \textbf{(7) \#~tokens (NT)} and \textbf{(8) \#~sentences (NS)} of ICL responses, as these format metrics can significantly impact model performance \citep{fan-etal-2018-controllable}. 
\emph{For each metric, we select the demonstrations having the same score and evaluate whether the ICL-generated responses maintain that score}. 

Our experiments are performed on 100 random \textbf{SAMSum} samples \citep{gliwa-etal-2019-samsum} for each metric. We use \textbf{ChatGPT (gpt-3.5-turbo-1106)} \citep{openai2022chatgpt} with \textbf{Self-consistency} \citep{wang2022self} to evaluate metrics (1)--(6) on a scale of 1--5, as it is an effective evaluator \citep{wang-etal-2023-chatgpt}. NLTK \citep{bird-loper-2004-nltk} assesses metrics (7)--(8). For ICL experiments, we examine \llong{five instruct and non-instruct models: \textbf{Mistral-7B-v0.3} \citep{jiang2023mistral}, \textbf{Llama-3.1-8B} \citep{dubey2024thel3}, \textbf{Qwen-2.5-7B} \citep{qwen25}, ans \textbf{Mistral-7B-it-v.02} \citep{jiang2023mistral} and \textbf{Llama-3.1-8B-it} \citep{dubey2024thel3}}.
For metrics (1)--(6), we select demonstrations having a perfect score of 5, and for metrics (7)--(8) having 17 response tokens, spanning 2 sentences. 
For each metric, we further examine whether a \textbf{simple guideline}: ``The output must maintain...\{property\}.'' 
({w/ guideline}) can help \llong{instruct models (Mistral)} maintain that property better during generation.

\paragraph{Findings.} 
We present the $k=5$ demonstration results \llong{in \Cref{tab:icl-motivation-results} and the case of Mistral-7B-it-v.02 in \Cref{fig:more-demos-does-not-help}} with metrics (1)--(7)\footnote{We also tested demonstration counts of $k={3,10}$, all follow similar trends,
see Appendix C.3's \Cref{fig:more-demos-does-not-help-full-results}.}. We derive three surprising findings. Firstly, the \llong{ICL models} do not achieve a 100\% score of $5$ on any metric \llong{and instruct models generally outperform non-instruct models}. \llong{The} highest percentage of score $5$ on average is on COH and REL, where \llong{ICL models} already excel while for critical summarization metrics such as INF and COV, they achieve only up to 20\% to 40\%. Notably, although all demonstrations contain 17 output tokens, fewer than 5\% answers achieve this property. Secondly, increasing \#~demonstrations does not rectify this issue; the same trends persist across 3, 5, and 10. Finally, by adding a simple guideline shown in Appx.\Cref{fig:more-demos-does-not-help}, the percentages of answers maintaining the metrics are mostly improved, especially (7) and (8), verifying that adding guidelines is indeed helpful for \llong{instruct models} to maintain these properties. Without ICL (and without instruction in our consideration), the model is entirely unable to solve the task.

\paragraph{Theoretical intuitions.} Our theoretical intuitions explaining the above observations are provided in \Cref{ssec:theoretical-analysis}. In summary, we prove that when an LLM does not fully capture the true task distribution ($P_{\mathcal{M}} \neq P_{T}$), demonstrations cannot recover the true task distribution in the limit, causing certain task-specific language and format properties may not be preserved in generation. We term this as the \textbf{text property transfer (PT)} problem. To address this, we define a \emph{text property task} as a reformulation of the original task, where responses are mapped via a property-specific metric. We hypothesize that the original task can be approximated by optimizing a set of well-chosen text property tasks, allowing the model to better align with desired text characteristics and improve generation quality.

\begin{figure*}[t]
\centering
\includegraphics[width=1\linewidth, trim={0cm 0cm 0cm 0cm},clip]{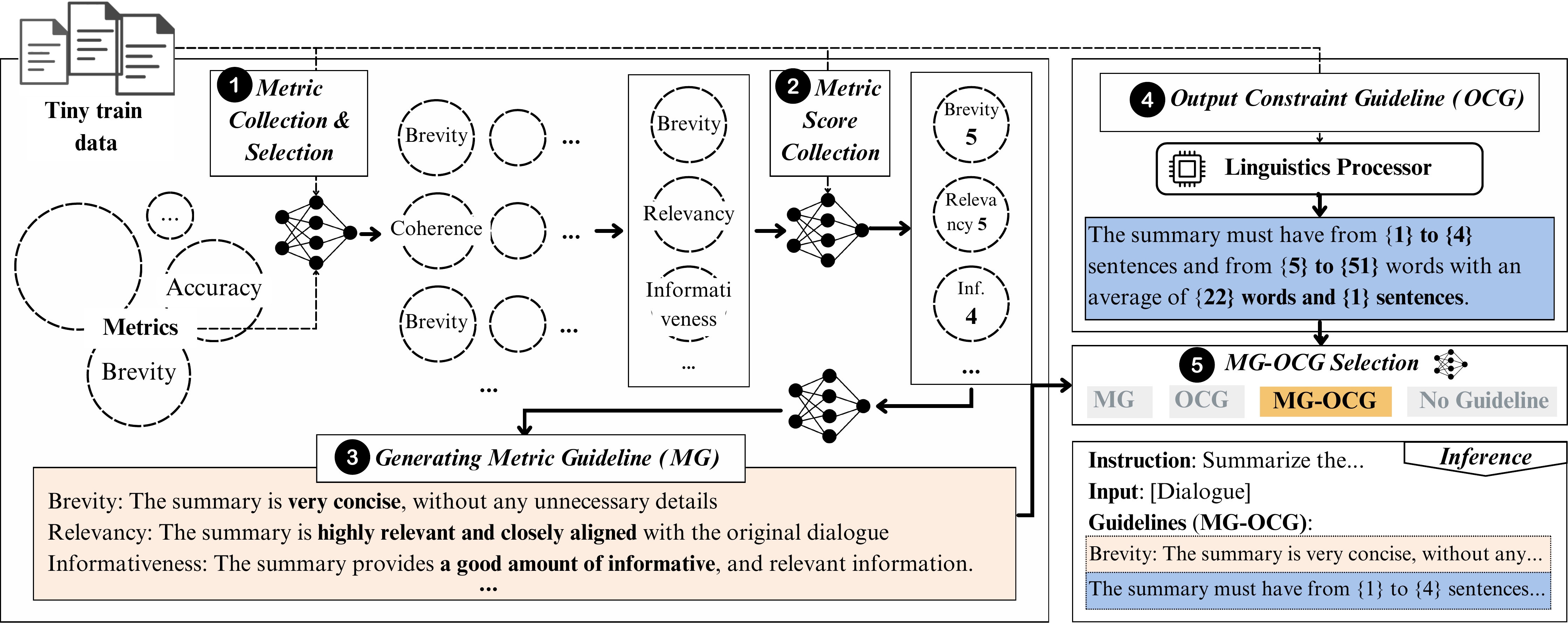}
\caption{\small{Overview of \model{}. Orange and blue boxes denote the learned metric guideline and output constraint guideline.} 
}
\label{fig:overview-framework}
\end{figure*}

\section{\model{}: An Efficient Guideline Generation Algorithm} \label{sec:method}

\paragraph{Motivations.} 
As we have seen, providing textual \textbf{guidelines} instructing LLMs to optimize certain text property metrics can enhance them on responses, 
possibly because LLMs are optimizers \citep{yang2024large}. We propose \model{} (\Cref{fig:overview-framework} and \Cref{algo:longguide}), 
an algorithm that efficiently generates guidelines for LLMs to optimize self-evaluated text properties during generation. Specifically, Steps 1--3 focus on generating the Metric Guideline (MG) capturing the intrinsic language properties of the task via reference-free metrics. In parallel, Step 4 analyzes the answer format of the task and translates it to Output Constraint Guideline (OCG). The best combination of MG and OCG is selected for inference (Step 5). To ensure \model{}'s generalizability to new tasks, we assume access to at most $50$ training samples: $D^{train} = \{(x^{t}_i, y^{t}_i)\}_{i=1}^{n}$. 

\paragraph{Step 1: Metric Collection \& Selection.} To learn a task's language properties, this step reasons to select appropriate language evaluation metrics for self-evaluation. For this purpose, we first construct a pool of evaluation metrics, $S$, applicable to any text generation task. $S$ consists of $27$ distinct metrics from $4$ main sources (Appx.-\Cref{tab:evaluation-metrics} for details). Specifically, we collect $3$ metrics from ABC's of Communication \citep{wagner1963abc}, $12$ metrics from \citep{ fu2023gptscore} for dialogue generation, summarization, data2text generation, and machine translation, and propose $12$ more metrics for a broader evaluation coverage. We do not collect LM-based metrics, such as FactScore \citep{min-etal-2023-factscore}, because it is challenging for LLMs to define and self-evaluate them. Additionally, we do not gather definitions of collected metrics, as their interpretations may vary across different tasks.

With $D^{train}$ and $S$, we then perform $K$ iterations to select the metrics. At each iteration, we randomly sample a batch of data from $D^{train}$ and instruct $\mathcal{M}$ to generate the top-5 most important metrics in $S$ for evaluating batch data properties via Chain-of-Thought prompting \citep{wei2022chain}.
We implement the top-5 constraint to avoid excessive metrics being selected. The final set of selected metrics, denoted by $M$, consists of the metrics chosen across all iterations \emph{sorted in alphabetic order}. 

\paragraph{Step 2: Metric Score Collection via Self-evaluation.} 
This step focuses on evaluating the selected metrics from $M$ on $D^{train}$ to capture the task properties. Motivated by prior studies \citep{wang-etal-2023-chatgpt,ren2023self}, we utilize $\mathcal{M}$ to score the metrics on a scale of 1--5. Specifically, for each train sample, $\mathcal{M}$ scores its ground-truth answer on all $\mathcal{M}$'s metrics via Self-Consistency \citep{wang2022self}. The final metrics' scores, denoted as $scores_{M}$, are the average of scores over all train samples.
Note that we separate this step from Step 1's metric selection because we want to evaluate each chosen metric on $D^{train}$ instead of the samples that led $\mathcal{M}$ to select it.

\begin{table*}
\centering
\footnotesize
\scalebox{.87}{
\begin{tabular}{l|cccc|cccc|cccc}
\toprule
 & & \textbf{SAMSum} & & & & \textbf{CNN} & & & & \textbf{SWiPE} &  \\
\midrule
\textbf{Method} & \textbf{R-L\textcolor[RGB]{0,0,0}{$\uparrow$}} & \textbf{B-1\textcolor[RGB]{0,0,0}{$\uparrow$}} & \textbf{BS\textcolor[RGB]{0,0,0}{$\uparrow$}} & \textbf{Avg.JS\textcolor[RGB]{0,0,0}{$\downarrow$}} & \textbf{R-L\textcolor[RGB]{0,0,0}{$\uparrow$}} & \textbf{B-1\textcolor[RGB]{0,0,0}{$\uparrow$}} & \textbf{BS\textcolor[RGB]{0,0,0}{$\uparrow$}} & \textbf{Avg.JS\textcolor[RGB]{0,0,0}{$\downarrow$}} & \textbf{R-L\textcolor[RGB]{0,0,0}{$\uparrow$}} & \textbf{B-1\textcolor[RGB]{0,0,0}{$\uparrow$}} & \textbf{BS\textcolor[RGB]{0,0,0}{$\uparrow$}} & \textbf{Avg.JS\textcolor[RGB]{0,0,0}{$\downarrow$}} \\
\midrule
Zero-shot (ZS) & 22.20  & 20.05 & 58.98 & 0.1014  &  19.23 & 20.43 & 60.59 &  0.1262 & 36.60 & 39.01 & \textbf{71.18} & 0.0565 \\
\cdashline{2-10}
\emph{+ OCG} & 27.55 & 28.64 & 60.38 & 0.0402 & \textbf{22.46} & \textbf{27.82} & \textbf{61.37} & \textbf{0.0718} & 32.48 & 32.88 & {67.32} & 0.0650 \\
\emph{+ MG} & 27.81 & \textbf{28.81} & 60.06 & 0.0388  & 18.35 & 19.66 & {59.79} & 0.1413 & \textbf{38.21} & \textbf{40.83} & {70.87} & \textbf{0.0550} \\
\emph{+ MG-OCG} & \textbf{28.35} & 28.79 & \textbf{60.66} & \textbf{0.0375} & 22.05 & 26.97 & 61.18 & 0.0789 & 35.47 & 36.95 & {68.77} & 0.0554 \\
\cdashline{1-13}
\textbf{\emph{+ \model{}}} & \textbf{28.35} & 28.79 & \textbf{60.66} & \textbf{0.0375} & \textbf{22.46} & \textbf{27.82} & \textbf{61.37} & \textbf{0.0718} & \textbf{38.21} & \textbf{40.83} & {70.87} & \textbf{0.0550} \\
\cmidrule{1-13}
Few-shot (FS)  & 27.13 & 27.21 & 61.70 & 0.0502 & 17.56 & 20.55 & 57.74 & 0.0844 & 39.47 & 39.76 & 70.56 & 0.0469 \\
\cdashline{1-13}
\emph{+ OCG} & 27.84 & 29.91 & 61.08 & 0.0336 & 15.20 & 17.58 & 58.12 & 0.0922 & 29.54 & 30.32 & 68.82 & 0.0596 \\
\emph{+ MG} & 27.50 & 30.15 & 62.24 & 0.0352 & 18.13 & 20.94 & 57.89 & 0.0830 &  \textbf{41.36} & \textbf{41.22} & \textbf{71.14} & \textbf{0.0450} \\
\emph{+ MG-OCG} & \textbf{30.65} & \textbf{31.72} & \textbf{62.73} & \textbf{0.0318} & \textbf{19.19} & \textbf{22.30} & \textbf{57.95} & \textbf{0.0814} & 38.56 & 37.87 & 68.54 & 0.0529 \\
\cdashline{1-13}
\textbf{\emph{+ \model{}}} & \textbf{30.65} & \textbf{31.72} & \textbf{62.73} & \textbf{0.0318} & \textbf{19.19} & \textbf{22.30} & \textbf{57.95} & \textbf{0.0814} & \textbf{41.36} & \textbf{41.22} & \textbf{71.14} & \textbf{0.0450} \\
\bottomrule
\end{tabular}}
\caption{
\small{Mistral performance verifying \model{} mitigates the text property transfer (PT) problem (\Cref{sec:demos-anayn}): (1) the trends of ROUGE-L (R-L), BLEU-1 (B-1), BERTScore (BS), and Jensen–Shannon divergence (Avg. JS) show strong correlations, supporting our hypothesis; (2) \model{} substantially enhances Avg. JS scores, thereby mitigating the PT problem.}}
\label{table:does-longguide-address-PL-problem}
\end{table*}

\paragraph{Step 3: Generating {M}etric {G}uideline (MG).}  This step aims to generate a textual metric guideline (MG) that guides $\mathcal{M}$ to align generation outputs with task-specific properties from $scores_{M}$. MG is formed by concatenating metrics' definitions generated by $\mathcal{M}$ and tailored by $scores_{M}$ via the LLM instruction ``Based on these scores on a scale of 5...define the expected quality of the output for each
metric in natural language''. We use these moderated definitions instead of raw $scores_{M}$ because LLMs better capture contextual nuances through descriptions rather than numerical scores \citep{singh2024tokenization}. \Cref{fig:overview-framework} illustrates an instance where ``Informativeness'' in the task ``dialogue sum.'' achieving $4/5$ score from Step 2 is defined as {``...\underline{good} amt. of informative...''}.

\paragraph{Step 4: {O}utput {C}onstraint {G}uideline (OCG).} 
Research on controllable generation has extensively proposed constraints including ones on the length, which are broadly applicable, as well as linguistic or keyword, which are more task-specific \citep{fan-etal-2018-controllable,he-etal-2022-ctrlsum}. In this step, we aim to establish a robust set of output constraints that apply universally to long-form generation tasks. We focus on six key constraints related to two distributions: the number of sentences and tokens in ground-truth answers. These constraints include minimum, maximum, and average counts, serving as basic statistics for length and expected values. The Output Constraint Guideline (OCG) instructs $\mathcal{M}$ to adhere to these statistics during generation.

\paragraph{Step 5: MG--OCG selection.}

Models have varying inherent knowledge per task, resulting in different improvements gained by using MG and OCG (\Cref{table:main-ablation-results}). This step determines the optimal MG and OCG combination by evaluating model performance on $D^{train}$ across configurations. The best-performing guideline configuration is then selected as the final output of \model{}.

\section{Experiments} \label{sec:experimentation}

\paragraph{Benchmarks.} We benchmark \model{} on seven widely evaluated long-form generation tasks from four main categories, \emph{summarization, text simplification, machine translation and generation}, \llong{and one real-life chat LLM benchmark}. These tasks are \textbf{SAMSum} \citep{gliwa-etal-2019-samsum}, \textbf{CNN/Daily Mail (3.0.0)} \citep{see-etal-2017-get} and \textbf{XL-SUM} \citep{hasan-etal-2021-xl} for summarization, \textbf{SWiPE} \citep{laban-etal-2023-swipe} for text simplification, \textbf{IWSLT-2017 en-ja} \citep{cettolo-etal-2017-overview} for machine translation, \textbf{Synthetic-Persona-Chat} \citep{jandaghi2023faithful} for dialogue generation,  \textbf{CommonGen-Challenge} \citep{lin-etal-2020-commongen} for data-to-text generation, \llong{and (a subset of) \textbf{AlpacaEval2} \citep{dubois2024lengthcontrolled}}. We also benchmark the \textbf{reasoning tasks} in \Cref{appx:longguide-in-reasoning}. See \Cref{Appx:baselines-details} for details.

\paragraph{Baselines and evaluations.} Since \model{} is the first method to self-learn guidelines as additional instructions for generation, we compare it with the \textbf{zero-/few-shot prompting} baselines in this section, \llong{and \textbf{many-shot prompting} in \Cref{ssec:extra-baselines-more-shots}}.  We also evaluate it against \llong{three} of the strongest prompt optimization algorithms to date: \textbf{APO} \citep{pryzant-etal-2023-automatic} in this section, and \llong{\textbf{EvolPrompt} \citep{guo2024connecting}} and \textbf{adv-ICL} \citep{do2024prompt} in \Cref{sec:longguide-can-be-combined}, both of which optimize the input prompt on the $D^{train}$. We also compare \model{} with \textbf{``General Guidelines''} in \Cref{ssec:generalize-from-demos} where we ask the models to reason over demonstrations to generate task guidelines. For models, we empirically examine both strong open- and closed-source LLMs: \textbf{Mistral-7B-it v0.2} \citep{jiang2023mistral} as an open-source model and \textbf{ChatGPT (gpt-3.5-turbo-1106)} \citep{openai2022chatgpt} as a closed-source model. For evaluations, we use \textbf{ROUGE-L} \citep{lin-2004-rouge} (recall-based) following \citet{bai2023longbench} (also for \model{}'s Step 5), and \llong{\textbf{GPT-4o-Judge} \citep{openai2024gpt4o} as our main evaluation metrics}. For \llong{{GPT-4o-Judge}, we evaluate how aligned the generated answer is with the reference answer and its quality on five criteria: \emph{(i) Format consistency}; \emph{(ii) Content completeness}; \emph{(iii) Factuality}; \emph{(iv) Style adherence}; \emph{(v) Generation quality} on a scale of 1--10, following \citet{zheng2023judging}  (see \Cref{sec:prompt-for-gpt4-judge} for full prompt)}. We also report \textbf{BLEU-1} \citep{papineni-etal-2002-bleu} (precision-based), \textbf{BERTScore} \citep{Zhang2020BERTScore} (meaning-based), and \textbf{Human evaluation} verifying the metric optimization and generation quality in \Cref{ssec:human-eval}. The results are averaged over \textbf{three runs}, with 95\% CI of the t-test.

\subsection{Findings} \label{ssec:main-findings}


\begin{table*}
\centering
\footnotesize
\scalebox{0.87}{
\begin{tabular}{cl|ccc|c|c|c|c}
\toprule
 & &  & \textbf{Sum.} & & \textbf{Simplification} & \textbf{Translation} & \textbf{Dialogue Gen.}  &\textbf{Table2Text} \\
\midrule
& \textbf{Method} & \textbf{SAMSum} & \textbf{CNN (3.0.0)} & \textbf{XL-Sum} & \textbf{SWiPE} & \textbf{IWSLT17 en-ja} & \textbf{Syn. Persona} & \textbf{Comm.-Chall.} \\
\midrule
& \textbf{\#shots (ran.)} &  3 & 3 & 5 & 3 & 3 &  5 & 5 \\
\midrule
\multirow{9}*{\begin{tabular}[c]{@{}l@{}}{\rotatebox{90}{\textbf{Mistral-it (0.2)}}}
\end{tabular}} 
& Zero-shot &  22.20 / \llong{7.43} & 19.23 / \llong{7.38} & 9.19 / \llong{5.96} &  36.60 / \llong{7.21} &  13.12 / \llong{2.82} &  12.76 / \llong{2.68} &  10.12 / \llong{5.14} \\
&+ APO & 23.77 / \llong{7.31} & 19.53 / \llong{7.40} & 12.06 / \llong{5.85} & 36.92 / \llong{7.21}  & 14.45 / \llong{2.91} & 10.66 / \llong{2.41} & 11.21 / \llong{4.68}   \\
\cdashline{2-9}
&\textbf{\emph{+ \model{}}} & \textbf{28.35} / \textbf{\llong{7.73}} & \textbf{22.46} / \textbf{\llong{7.45}} & \textbf{14.38} / \textbf{\llong{6.29}} & \textbf{38.21} / \textbf{\llong{7.32}} & \textbf{16.53} / \textbf{\llong{3.45}} & \textbf{14.69} / \textbf{\llong{4.45}} & \textbf{25.20} / \textbf{\llong{6.81}}  \\
& \textcolor{gray}{\emph{\% gain (+)}} & \textcolor[RGB]{0,80,71}{$6.15$} / \textcolor[RGB]{0,80,71}{$0.30$} & \textcolor[RGB]{0,80,71}{$3.23$} / \textcolor[RGB]{0,80,71}{$0.07$}  & \textcolor[RGB]{0,80,71}{$5.19$} / \textcolor[RGB]{0,80,71}{$0.33$} & \textcolor[RGB]{0,80,71}{$1.61$} / \textcolor[RGB]{0,80,71}{$0.11$} & \textcolor[RGB]{0,80,71}{$3.41$} / \textcolor[RGB]{0,80,71}{$0.63$} & \textcolor[RGB]{0,80,71}{$1.93$} / \textcolor[RGB]{0,80,71}{$1.77$} & \textcolor[RGB]{0,80,71}{$15.08$} / \textcolor[RGB]{0,80,71}{$1.67$} \\
\cmidrule{2-9}
&Few-shot  & 27.13 / \llong{7.66} & 17.56 / \llong{5.84} & 9.79 / \llong{4.46} & 39.47 / {\llong{7.12}} & 12.69 / \llong{2.66} &  3.56 / \llong{1.00} & 3.98 / \llong{1.34} \\
&+ APO & 26.23 / \llong{7.44} & 18.18 / \llong{5.89} & 11.99 / \llong{4.55} & 39.55 / \llong{7.11} & 14.08 / \llong{2.92} & 4.26 / \llong{1.05} & 5.45 / \llong{2.05} \\
\cdashline{2-9}
& \textbf{\emph{+ \model{}}} & \textbf{30.65} / \textbf{\llong{7.72}} & \textbf{19.19} / \textbf{\llong{5.99}} & \textbf{15.23} / \textbf{\llong{5.06}} & \textbf{41.36} / \textbf{\llong{7.24}} & \textbf{16.62} / \textbf{\llong{3.40}} & \textbf{5.25} / \textbf{\llong{3.93}} & \textbf{25.05} / \textbf{\llong{6.65}} \\
& \textcolor{gray}{\emph{\% gain (+)}} & \textcolor[RGB]{0,80,71}{$3.52$} / \textcolor[RGB]{0,80,71}{$0.06$} & \textcolor[RGB]{0,80,71}{$1.63$} / \textcolor[RGB]{0,80,71}{$0.15$} & \textcolor[RGB]{0,80,71}{$5.44$} / \textcolor[RGB]{0,80,71}{$0.40$}  & \textcolor[RGB]{0,80,71}{$1.89$} / \textcolor[RGB]{0,80,71}{$0.12$} & \textcolor[RGB]{0,80,71}{$3.66$} / \textcolor[RGB]{0,80,71}{$0.74$} & \textcolor[RGB]{0,80,71}{$1.69$} / \textcolor[RGB]{0,80,71}{$2.93$} & \textcolor[RGB]{0,80,71}{$21.07$} / \textcolor[RGB]{0,80,71}{$5.31$} \\
\midrule
\multirow{7}*{\begin{tabular}[c]{@{}l@{}} {\rotatebox{90}{\textbf{ChatGPT}}} \\
\end{tabular}} 
& Zero-shot &  23.83 / \llong{7.43}  & 20.12 / \llong{7.44} & 10.80 / \llong{5.96} &  45.09 / \llong{7.28} &  36.13 / \llong{7.62} &  19.46 / \llong{6.04} & 24.21 / \llong{6.53} \\
&+ APO & 25.05 / \llong{7.45} & 20.34 / \llong{7.39} & 12.19 / \llong{6.07} & \textbf{46.32} / \textbf{\llong{7.51}} & 37.74 / \llong{7.44} & 19.91 / \llong{6.12} & 23.63 / \llong{6.53}\\
\cdashline{2-9}
& \textbf{\emph{+ \model{}}} & \textbf{30.47} / \textbf{\llong{7.59}} & \textbf{22.19} /  \textbf{\llong{7.67}} & \textbf{20.93} / \textbf{\llong{6.36}} & 45.09 / {\llong{7.28}} & \textbf{41.22} / \textbf{\llong{8.11}} & \textbf{22.98} / \textbf{\llong{6.41}} & \textbf{34.41} / \textbf{\llong{7.23}} \\
& \textcolor{gray}{\emph{\% gain (+)}} & \textcolor[RGB]{0,80,71}{$6.64$} / \textcolor[RGB]{0,80,71}{$0.16$}  & \textcolor[RGB]{0,80,71}{$2.07$} / \textcolor[RGB]{0,80,71}{$0.23$} & \textcolor[RGB]{0,80,71}{$10.13$} / \textcolor[RGB]{0,80,71}{$0.40$} & \textcolor[RGB]{0,80,71}{$0.00$} / \textcolor[RGB]{0,80,71}{$0.00$} & \textcolor[RGB]{0,80,71}{$5.09$} / \textcolor[RGB]{0,80,71}{$0.49$} & \textcolor[RGB]{0,80,71}{$3.52$} / \textcolor[RGB]{0,80,71}{$0.37$} & \textcolor[RGB]{0,80,71}{$10.20$} / \textcolor[RGB]{0,80,71}{$0.70$} \\
\cmidrule{2-9}
&Few-shot & 22.21 / \llong{7.32}  & 14.51 / \llong{4.38} &  11.42 / \llong{5.95} & 33.72 / \llong{5.07} & 31.93 / \llong{7.25} & 16.10 / \llong{4.67} & 22.08 / \llong{4.19} \\
&+ APO & 24.22 / \llong{7.28} & 15.20 / \llong{4.01} & 14.07 / \llong{6.19} & 34.46 / \llong{5.13} & 33.72 / \llong{7.31} & 17.68 / \llong{4.55} & 25.09 / \llong{6.12}\\
\cdashline{2-9}
& \textbf{\emph{+ \model{}}} & \textbf{31.46} / \textbf{\llong{7.72}}  & \textbf{18.17} / \textbf{\llong{4.42}} & \textbf{19.95} / \textbf{\llong{6.36}}  & \textbf{37.60} / \textbf{\llong{5.25}} & \textbf{38.43} / \textbf{\llong{7.91}} & \textbf{22.36} / \textbf{\llong{5.26}} & \textbf{38.21} / \textbf{\llong{7.21}} \\
& \textcolor{gray}{\emph{\% gain (+)}} & \textcolor[RGB]{0,80,71}{$9.25$} / \textcolor[RGB]{0,80,71}{$0.40$} & \textcolor[RGB]{0,80,71}{$3.66$} / \textcolor[RGB]{0,80,71}{$0.04$} & \textcolor[RGB]{0,80,71}{$8.53$} / \textcolor[RGB]{0,80,71}{$0.41$} & \textcolor[RGB]{0,80,71}{$3.88$} / \textcolor[RGB]{0,80,71}{$0.18$} & \textcolor[RGB]{0,80,71}{$6.50$} / \textcolor[RGB]{0,80,71}{$0.66$} & \textcolor[RGB]{0,80,71}{$6.53$} / \textcolor[RGB]{0,80,71}{$0.59$} & \textcolor[RGB]{0,80,71}{$16.13$} / \textcolor[RGB]{0,80,71}{$3.02$} \\
\bottomrule
\end{tabular}}
\caption{
\small{{ROUGE-L / \llong{GPT-4o-Judge}} results on seven long-form generation tasks. \model{} remarkably outperforms baselines on most tasks and substantially enhances LLMs. \llong{BLUE-1 scores are reported in Appx.-\ref{table:supplement-main-experimental-results}.}}}
\label{table:main-experimental-results}
\end{table*}

\paragraph{\model{} enhances PT which correlates with improved performance.}
\model{} effectively addresses the PT problem identified in \Cref{sec:demos-anayn}. Our experiments are conducted on $3$ benchmarks SAMSum, CNN, and SWiPE with Mistral under the zero-shot and few-shot settings. For each task, we first obtain the set of selected text properties from \model{} that the model needs to optimize. We then measure the average of Jensen-Shannon divergence \citep{lin1991divergence} between their score distributions (judged by ChatGPT) between the generated answers and the ground truth answers, across all selected properties, denoted as $Avg. JS$. 

The results are presented in \Cref{table:does-longguide-address-PL-problem}. \model{} significantly lowers the $Avg. JS$ scores compared to the baselines, demonstrating the effectiveness of guidelines for enhancing property transfer. Furthermore, our findings corroborate our hypothesis: across benchmarks, $Avg. JS$  exhibits moderate to strong positive correlations with the performance metrics (ROUGE-L, BLEU-1, BERTScore) measured by the Pearson correlation coefficient \citep{pearson1895vii} (Appx.-\Cref{fig:pearson-correlation}). In \Cref{appdx:js-components}, we present the density plots for all metrics measured on the results with and without \model{}.

\begin{figure}
\centering
\includegraphics[width=1\linewidth, trim={0cm 0cm 0cm 0cm},clip]{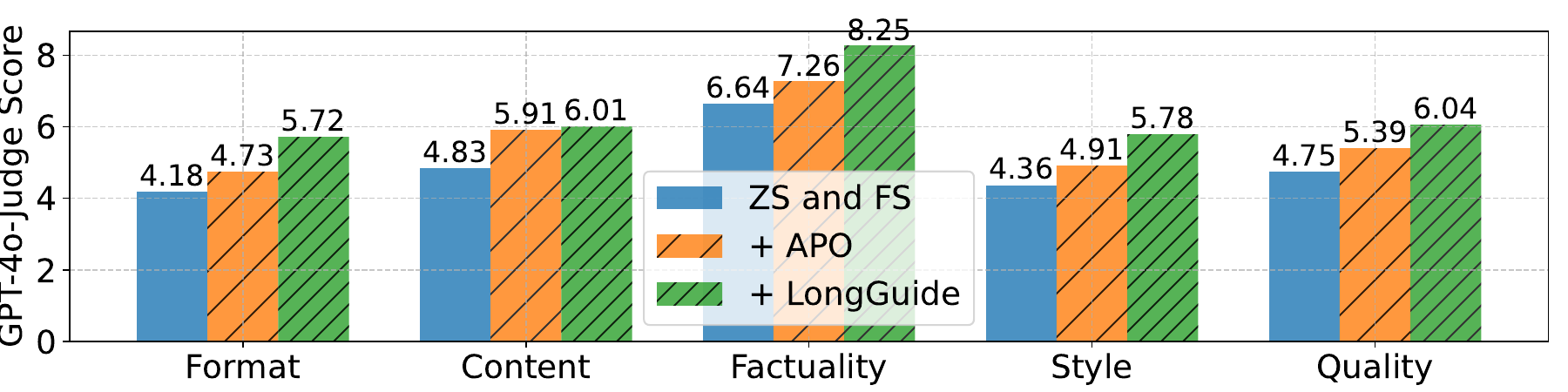}
\vspace{-7mm}
\caption{\small{GPT-4-Judge scores over criteria.}}
\vspace{-3mm}
\label{fig:gpt-judge-criteria}
\end{figure}

\Cref{table:main-experimental-results} details our main experiment results on downstream tasks. Firstly, for baselines, zero-shot performance is interestingly higher than the few-shot for both models, and the gaps are huge for Synthetic Persona and CommonGen-Challenge. We hypothesize that models were partly exposed to task data during training, causing few-shot demonstrations to push the prompts out of distribution, leading to frequent refusals to answer. Meanwhile, \model{} helps models overcome this issue. Secondly, \model{} substantially improves zero- and few-shot baselines by \llong{6\% on ROUGE-L and 0.8 on GPT-4o-Judge} on average: improvement for few-shot prompting is surprisingly higher than in zero-shot, possibly because improving a stronger baseline is harder than a weaker one. Notably, \model{} outperforms APO in most benchmarks, especially under the zero-shot setting, demonstrating that our strategy of optimizing text property tasks is markedly more effective than APO optimizing only ROUGE-L on limited data. Thirdly, we observe that \model{} achieves the highest improvements on CommonGen-Challenge with \llong{$15.62\%$} and lowest on SWiPE with \llong{$1.85\%$ on ROUGE-L}. These improvements are mainly because the answers generated by the baselines are often far longer than the ground truth. \model{} rectifies this issue by controlling the output length and quality, leading to substantial performance gains. Fourthly, among the two models, \model{} interestingly improves Mistral by $5.39\%$, while ChatGPT, regarded as stronger, improves by a larger margin, $6.58\%$. This suggests that \model{} has the potential to benefit stronger models in the future. 
\llong{Among five GPT-4o-Judge criteria in \Cref{fig:gpt-judge-criteria}, \model{} notably improves Format, Style, and Factuality, confirming its effectiveness in aligning model generation with ground-truth distributions. Finally, the significant gains in Quality, together with the ROUGE-L scores from \Cref{table:main-experimental-results} further verify that \model{} also strongly enhances the generation quality.}

\begin{table*}
\centering
\footnotesize
\scalebox{.83}{
\begin{tabular}{cl|ccc|c|c|c|c}
\toprule
 & \textbf{Method} & \textbf{SAMSum} & \textbf{CNN (3.0.0)} & \textbf{XL-Sum} & \textbf{SWiPE} & \textbf{IWSLT17 en-ja} & \textbf{Synthetic Persona} & \textbf{CommGen-Chall.} \\
\midrule
\multirow{11}*{\begin{tabular}[c]{@{}l@{}}{\rotatebox{90}{\textbf{Mistral-7B-it (0.2)}}}
\end{tabular}} 
& Zero-shot (ZS) &  22.20$_{\pm 0.43}$  &  19.23$_{\pm 0.34}$ &  9.19$_{\pm 0.03}$ &   36.60$_{\pm 0.59}$ &  13.12$_{\pm 1.39}$ &  12.76$_{\pm 1.54}$ &  10.12$_{\pm 0.02}$  \\
\cdashline{2-9}
& \emph{+ OCG} & 27.55$_{\pm 0.98}$\textcolor[RGB]{0,80,71}{$\uparrow$} & \textbf{22.46}$_{\pm 0.64}$\textcolor[RGB]{0,80,71}{$\uparrow$} & \textbf{14.38}$_{\pm 0.15}$\textcolor[RGB]{0,80,71}{$\uparrow$} & 32.48$_{\pm 1.91}$\textcolor[RGB]{250,0,0}{$\downarrow$} & \textbf{16.53}$_{\pm 0.59}$\textcolor[RGB]{0,80,71}{$\uparrow$} & 14.35$_{\pm 0.47}$\textcolor[RGB]{0,80,71}{$\uparrow$} & 24.16$_{\pm 0.11}$\textcolor[RGB]{0,80,71}{$\uparrow$} \\
& \emph{+ MG} & 27.81$_{\pm 1.17}$\textcolor[RGB]{0,80,71}{$\uparrow$} & 18.35$_{\pm 0.60}$\textcolor[RGB]{250,0,0}{$\downarrow$} & 9.37$_{\pm 0.25}$\textcolor[RGB]{0,80,71}{$\uparrow$} & \textbf{38.21}$_{\pm 1.72}$\textcolor[RGB]{0,80,71}{$\uparrow$} & 8.71$_{\pm 0.53}$\textcolor[RGB]{250,0,0}{$\downarrow$} & 12.53$_{\pm 0.58}$\textcolor[RGB]{250,0,0}{$\downarrow$} & 21.54$_{\pm 7.50}$\textcolor[RGB]{0,80,71}{$\uparrow$} \\
& \emph{+ MG-OCG} & \textbf{28.35}$_{\pm 1.66}$\textcolor[RGB]{0,80,71}{$\uparrow$} & 22.05$_{\pm 0.84}$\textcolor[RGB]{0,80,71}{$\uparrow$} & 13.64$_{\pm 0.38}$\textcolor[RGB]{0,80,71}{$\uparrow$} & 35.47$_{\pm 2.89}$\textcolor[RGB]{250,0,0}{$\downarrow$} & 15.76$_{\pm 1.85}$\textcolor[RGB]{0,80,71}{$\uparrow$} & \textbf{14.69}$_{\pm 1.08}$\textcolor[RGB]{0,80,71}{$\uparrow$} & \textbf{25.20}$_{\pm 1.89}$\textcolor[RGB]{0,80,71}{$\uparrow$}  \\
\cdashline{2-9}
& \textcolor{gray}{\emph{MG-OCG sel.}} & \textcolor{gray}{\emph{MG-OCG}} &  \textcolor{gray}{\emph{OCG}} &  \textcolor{gray}{\emph{OCG}} &  \textcolor{gray}{\emph{MG}} & \textcolor{gray}{\emph{OCG}} & \textcolor{gray}{\emph{MG-OCG}} & \textcolor{gray}{\emph{MG-OCG}} \\

\cmidrule{2-9}
& Few-shot (FS)  &  27.13$_{\pm 0.26}$ &  17.56$_{\pm 0.63}$ &  9.79$_{\pm 0.18}$ &  39.47$_{\pm 0.45}$ &  12.69$_{\pm 1.82}$ &  3.56$_{\pm 0.36}$ &  3.98$_{\pm 0.17}$ \\
\cdashline{2-9}
& \emph{+ OCG} & 27.84$_{\pm 0.88}$\textcolor[RGB]{0,80,71}{$\uparrow$} & 15.20$_{\pm 5.28}$\textcolor[RGB]{250,0,0}{$\downarrow$} & 12.22$_{\pm 1.19}$\textcolor[RGB]{0,80,71}{$\uparrow$} & 29.54$_{\pm 1.90}$\textcolor[RGB]{250,0,0}{$\downarrow$} & \textbf{16.62}$_{\pm 0.81}$\textcolor[RGB]{0,80,71}{$\uparrow$} & 5.06$_{\pm 1.05}$\textcolor[RGB]{0,80,71}{$\uparrow$} & \textbf{25.05}$_{\pm 0.76}$\textcolor[RGB]{0,80,71}{$\uparrow$} \\
& \emph{+ MG} & 27.50$_{\pm 2.08}$\textcolor[RGB]{0,80,71}{$\uparrow$} & 18.13$_{\pm 5.28}$\textcolor[RGB]{0,80,71}{$\uparrow$} & 11.80$_{\pm 2.06}$\textcolor[RGB]{0,80,71}{$\uparrow$} & \textbf{41.36}$_{\pm 1.37}$\textcolor[RGB]{0,80,71}{$\uparrow$} & 8.67$_{\pm 0.62}$\textcolor[RGB]{250,0,0}{$\downarrow$} & 4.32$_{\pm 0.39}$\textcolor[RGB]{0,80,71}{$\uparrow$} & 14.58$_{\pm 2.24}$\textcolor[RGB]{0,80,71}{$\uparrow$} \\
& \emph{+ MG-OCG} & \textbf{30.65}$_{\pm 0.88}$\textcolor[RGB]{0,80,71}{$\uparrow$} & \textbf{19.19}$_{\pm 0.49}$\textcolor[RGB]{0,80,71}{$\uparrow$} & \textbf{15.23}$_{\pm 0.33}$\textcolor[RGB]{0,80,71}{$\uparrow$} & 38.56$_{\pm 1.39}$\textcolor[RGB]{250,0,0}{$\downarrow$} & 15.83$_{\pm 0.95}$\textcolor[RGB]{0,80,71}{$\uparrow$} & \textbf{5.25}$_{\pm 0.94}$\textcolor[RGB]{0,80,71}{$\uparrow$} & 5.94$_{\pm 1.00}$\textcolor[RGB]{0,80,71}{$\uparrow$}  \\
\cdashline{2-9}
& \textcolor{gray}{\emph{MG-OCG Sel.}} & \textcolor{gray}{\emph{MG-OCG}} & \textcolor{gray}{\emph{MG-OCG}} & \textcolor{gray}{\emph{MG-OCG}} &  \textcolor{gray}{\emph{MG}} & \textcolor{gray}{\emph{OCG}} & \textcolor{gray}{\emph{MG-OCG}} & \textcolor{gray}{\emph{OCG}} \\

\midrule
\multirow{10}*{\begin{tabular}[c]{@{}l@{}} {\rotatebox{90}{\textbf{ChatGPT (1106)}}} \\
\end{tabular}} 
& Zero-shot (ZS) &  23.83$_{\pm 0.54}$ &  20.12$_{\pm 0.27}$ &  10.80$_{\pm 0.18}$ &  \textbf{45.09}$_{\pm 1.45}$ &  36.13$_{\pm 0.87}$ &  19.46$_{\pm 0.40}$ &  24.21$_{\pm 0.37}$ \\
\cdashline{2-9}
& \emph{+ OCG} & 29.19$_{\pm 0.77}$\textcolor[RGB]{0,80,71}{$\uparrow$} & \textbf{22.39}$_{\pm 0.82}$\textcolor[RGB]{0,80,71}{$\uparrow$} & \textbf{20.93}$_{\pm 0.52}$\textcolor[RGB]{0,80,71}{$\uparrow$} & 37.76$_{\pm 1.44}$\textcolor[RGB]{250,0,0}{$\downarrow$} & 38.86$_{\pm 1.11}$\textcolor[RGB]{0,80,71}{$\uparrow$} & \textbf{22.98}$_{\pm 2.65}$\textcolor[RGB]{0,80,71}{$\uparrow$} & \textbf{34.41}$_{\pm 1.01}$\textcolor[RGB]{0,80,71}{$\uparrow$} \\
& \emph{+ MG} & 25.38$_{\pm 0.79}$\textcolor[RGB]{0,80,71}{$\uparrow$} & 20.37$_{\pm 0.41}$\textcolor[RGB]{0,80,71}{$\uparrow$} & 10.42$_{\pm 1.15}$\textcolor[RGB]{250,0,0}{$\downarrow$} & 45.06$_{\pm 2.96}$\textcolor[RGB]{250,0,0}{$\downarrow$} & 37.88$_{\pm 2.42}$\textcolor[RGB]{0,80,71}{$\uparrow$} & 19.91$_{\pm 0.59}$\textcolor[RGB]{0,80,71}{$\uparrow$} & 17.23$_{\pm 2.57}$ \\
& \emph{+ MG-OCG} & \textbf{30.47}$_{\pm 1.57}$\textcolor[RGB]{0,80,71}{$\uparrow$} & \textbf{22.19}$_{\pm 0.65}$\textcolor[RGB]{0,80,71}{$\uparrow$} & 20.02$_{\pm 0.89}$\textcolor[RGB]{0,80,71}{$\uparrow$} & 41.38$_{\pm 4.91}$\textcolor[RGB]{250,0,0}{$\downarrow$} & \textbf{41.22}$_{\pm 0.46}$\textcolor[RGB]{0,80,71}{$\uparrow$} & 20.95$_{\pm 1.91}$\textcolor[RGB]{0,80,71}{$\uparrow$} & 31.57$_{\pm 0.99}$\textcolor[RGB]{0,80,71}{$\uparrow$}\\
\cdashline{2-9}
& \textcolor{gray}{\emph{MG-OCG Sel.}} & \textcolor{gray}{\emph{MG-OCG}}  & \textcolor{gray}{\emph{MG-OCG}} & \textcolor{gray}{\emph{OCG}} & \textcolor{gray}{\emph{ZS}} & \textcolor{gray}{\emph{MG-OCG}} &  \textcolor{gray}{\emph{MG-OCG}} & \textcolor{gray}{\emph{OCG}} \\
\cmidrule{2-9}
& Few-shot (FS)  &  22.21$_{\pm 2.35}$ &  14.51$_{\pm 0.80}$ &  11.42$_{\pm 0.13}$ &  33.72$_{\pm 2.61}$ &  31.93$_{\pm 1.88}$ &  16.10$_{\pm 2.61}$ &  22.08$_{\pm 0.63}$  \\
\cdashline{2-9}
& \emph{+ OCG} & 30.00$_{\pm 1.07}$\textcolor[RGB]{0,80,71}{$\uparrow$} & \textbf{18.17}$_{\pm 1.32}$\textcolor[RGB]{0,80,71}{$\uparrow$} & \textbf{19.95}$_{\pm 1.38}$\textcolor[RGB]{0,80,71}{$\uparrow$} & 16.68$_{\pm 1.29}$\textcolor[RGB]{250,0,0}{$\downarrow$} & 38.57$_{\pm 1.81}$\textcolor[RGB]{0,80,71}{$\uparrow$} & \textbf{22.36}$_{\pm 0.89}$\textcolor[RGB]{0,80,71}{$\uparrow$} & 38.12$_{\pm 1.99}$\textcolor[RGB]{0,80,71}{$\uparrow$} \\
& \emph{+ MG} & 29.43$_{\pm 0.83}$\textcolor[RGB]{0,80,71}{$\uparrow$}  & 15.45$_{\pm 2.16}$\textcolor[RGB]{0,80,71}{$\uparrow$} & 12.49$_{\pm 0.59}$\textcolor[RGB]{0,80,71}{$\uparrow$} & 19.36$_{\pm 1.40}$\textcolor[RGB]{250,0,0}{$\downarrow$} & \textbf{39.45}$_{\pm 3.55}$\textcolor[RGB]{0,80,71}{$\uparrow$} & 18.64$_{\pm 0.49}$\textcolor[RGB]{0,80,71}{$\uparrow$}  & 22.18$_{\pm 7.50}$\textcolor[RGB]{0,80,71}{$\uparrow$} \\
& \emph{+ MG-OCG} & \textbf{31.46}$_{\pm 1.34}$\textcolor[RGB]{0,80,71}{$\uparrow$} & 14.84$_{\pm 2.58}$\textcolor[RGB]{0,80,71}{$\uparrow$} & 18.58$_{\pm 0.44}$\textcolor[RGB]{0,80,71}{$\uparrow$} & \textbf{37.60}$_{\pm 2.85}$\textcolor[RGB]{0,80,71}{$\uparrow$} & 38.43$_{\pm 2.37}$\textcolor[RGB]{0,80,71}{$\uparrow$} & 19.47$_{\pm 1.20}$\textcolor[RGB]{0,80,71}{$\uparrow$} & \textbf{38.21}$_{\pm 3.70}$\textcolor[RGB]{0,80,71}{$\uparrow$} \\
\cdashline{2-9}
& \textcolor{gray}{\emph{MG-OCG Sel.}} & \textcolor{gray}{\emph{MG-OCG}} & \textcolor{gray}{\emph{OCG}} & \textcolor{gray}{\emph{OCG}} & \textcolor{gray}{\emph{MG-OCG}} & \textcolor{gray}{\emph{MG-OCG}} & \textcolor{gray}{\emph{OCG}} & \textcolor{gray}{\emph{MG-OCG}} \\
\bottomrule
\end{tabular}}
\caption{\small{ROUGE-L results with 95\% CI from t-test. The gains of \model{}'s components vary across different models and tasks.  The ``MG-OCG selection'' results are reported in Appx.-\Cref{table:CD-MG-selection-results}.}}
\label{table:main-ablation-results}
\end{table*}

\paragraph{Where do the enhancements come from?} 
To identify the primary source of performance gains, we present the results of LLMs with \model{}'s components in \Cref{table:main-ablation-results}. Firstly, MG-OCG combination ({w/ MG-OCG}) is the most useful for LLMs, observed to be the best $15$ times, followed by OCG ({w/ OCG}) with $10$, and MG ({w/ MG}) twice. While these statistics underscore the effectiveness of MG-OCG, OCG particularly proves itself highly effective in tasks such as summarization, translation, and table-to-text generation. Secondly, individual MG or OCG strengthens the prompting baselines, with OCG showing a slight edge. This is because while MG focuses on the language properties of answers, it does not directly control the output format, sometimes causing longer/shorter answers than the ground truths. Exceptionally, on SWiPE, OCG harms all models, whereas MG shows notably strong effectiveness with Mistral. Manual investigations reveal that ground-truth answers in SWiPE exhibit high variances in \#sentences and \#tokens which explains why OCG may not be effective for this benchmark. Thirdly, an interesting case is ChatGPT with few-shot prompting on SWiPE, where individual MG and OCG impair performance but their combination enhances it. This shows evidence that MG and OCG complement each other. As discussed above, due to the uneven nature of answers in SWiPE, using MG or OCG alone may not work well for multiple samples, as MG and OCG only provide expected statistics. However, combining them could enhance performance by allowing them to complement each other. A such complement SWiPE example is outlined in Appx.-\Cref{fig:out-of-distribution-example}.

\section{Discussion} \label{sec:discussion}

We address several key questions about the usefulness, applicability, and generalizability of \model{}. Additional properties are provided in \Cref{Appx:longguide-properties} and more method and attention analyses in \Cref{Appx:extra-analysis}. 

\subsection{Human Evaluation: Does \model{} Enhance Generation Quality?} \label{ssec:human-eval}

\begin{figure}[h]
\small 
\centering
\includegraphics[width=1\linewidth, trim={0cm 0cm 0cm 0cm},clip]{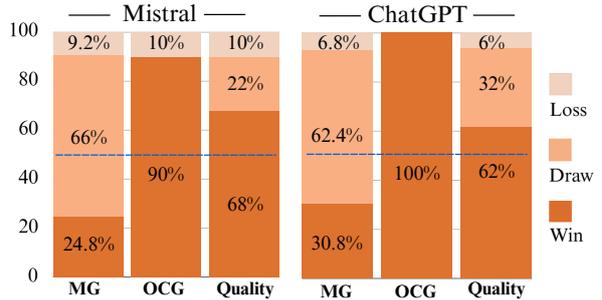}
\caption{\small{Win/Draw/Loss rates of w/ versus w/o \model{}.}}
\label{fig:human-eval}
\end{figure}

We perform a human evaluation to quantify whether \model{} helps LLMs optimize the selected metrics and enhance generation quality, as no automatic methods can address this need to date. For this purpose, we randomly select 50 zero-shot generated samples from the SAMSum and Syn. Persona (since MG-OCG is the best for these datasets, \Cref{table:main-ablation-results}). Three excellent English-native undergrads are hired to rate whether ZS + \model{} improves ZS on each of the selected MG and OCG metrics. Due to limited resources, we evaluate $5$ random MG metrics.

As shown in \Cref{fig:human-eval},
we notice that ZS + \model{} outperforms ZS on 27.8\% MG metrics on average, draws on 64.2\%, and loses on only 8\%. Specifically, among the MG metrics, ``Brevity'' shows the highest winning rate of $73\%$ while ``Relevance'' obtains the lowest winning rate of $12\%$, possibly because ZS models can already generate highly relevant outcomes. Meanwhile, on the OCG metrics, \model{} achieves a superior win of $95\%$ on average. Finally, regarding the generation quality, our annotators prefer \model{} output by up to $92\%$. These indicate that \model{} not only aligns the outputs with the ground truths but also enhances the generation quality. The fine-grained scores of MG metrics are provided in \Cref{ssec:human-evaluation-rating}, and annotator agreement (Krippendorff’s alpha \citep{krippendorff2022reliability}) is $\alpha$=68.9\%.


\subsection{\model{} Learns From Demonstrations To Boost ICL Performance} \label{ssec:generalize-from-demos}

\begin{figure}
\small 
\centering
\includegraphics[width=1\linewidth, trim={0cm 0cm 0cm 0cm},clip]{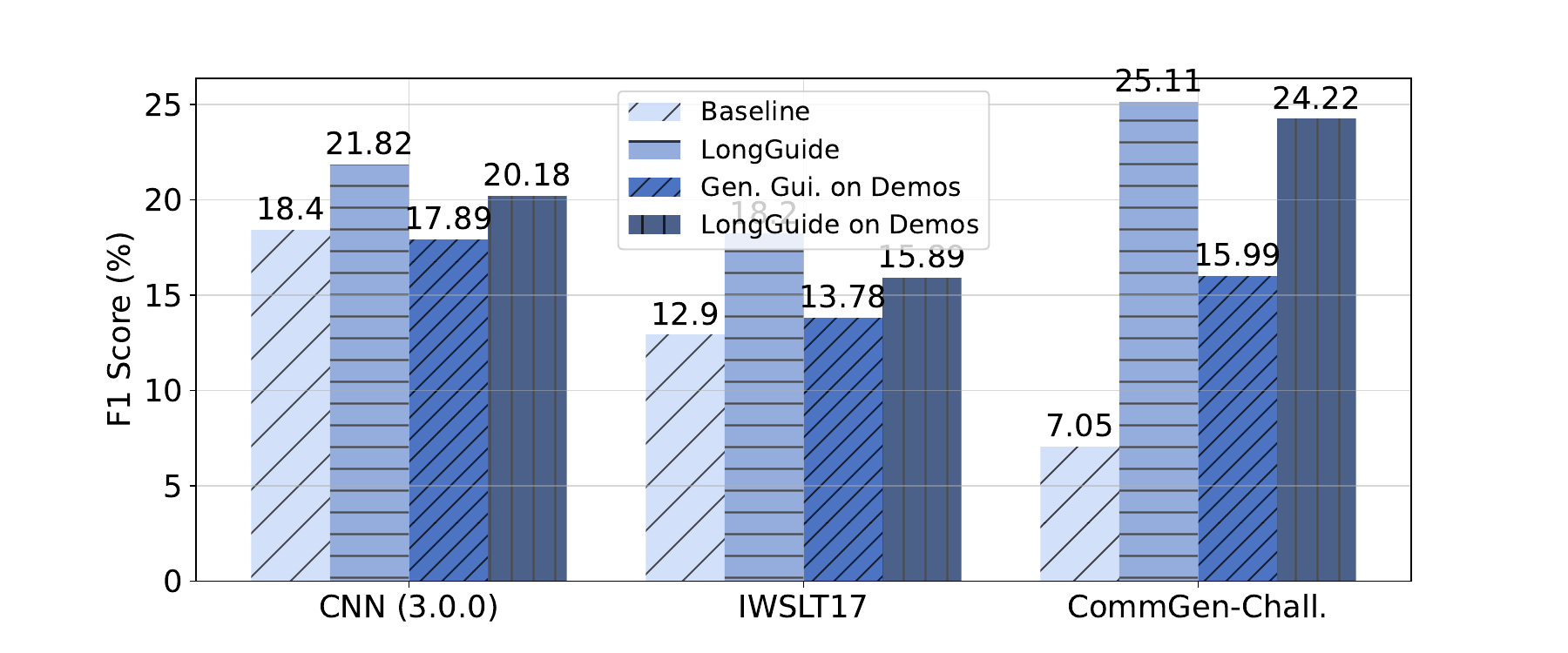}
\caption{\small{\model{} learned from demonstrations substantially enhances Mistral performance (ROUGE-L).}}
\label{fig:longguide-learns-from-demos}
\end{figure}

Here, we revisit the question posed in \Cref{sec:demos-anayn} and verify that \model{} learned from demonstrations substantially increases ICL performance. Our experiments using Mistral cover CNN, IWSLT17 en-ja, and CommGen-Chall datasets. Our experiments involve averaging the performance under zero- and few-shot settings. For \textbf{Baseline}, no guideline is utilized. For \textbf{\model{} on Demos}, we train \model{} on demonstrations used in \Cref{table:main-experimental-results}, in contrast to the $D^{train}$ for the case of \model{}. We add one more baseline, \textbf{General Guidelines (Gen. Gui.) on Demos}, where we ask the models to generate general task guidelines from demonstrations.

The results are summarized in \Cref{fig:longguide-learns-from-demos}, with details in Appx.-\Cref{tab:longpo-learns-from-demos}. Specifically, \model{} trained on $D^{train}$ outperforms it on demonstrations, suggesting its possible scalability with more training data. Moreover, while Gen. Gui. slightly worsens the Baseline on CNN, both \model{} and \model{} on Demos notably surpass the Baseline, and Gen. Gui., highlighting the effectiveness of \model{} in capturing task-specific properties, thereby boosting ICL performance.

\subsection{Ablation Studies of \model{}'s Steps}

\begin{table}[t]
\centering
\scalebox{.63}{
\begin{tabular}{lcccc}
\toprule 
\textbf{Methods} & \textbf{SAMSum} & \textbf{SWiPE} & \textbf{CommGen-Chall.}\\
\midrule
Zero-shot (ZS) & 22.20 / \llong{7.43} & 36.60 / \llong{7.21}  & 10.12 /  \llong{5.14} \\
\emph{\textbf{+ \model{}}} & \textbf{28.35} / \llong{\textbf{7.73}} & \textbf{38.21} / \llong{\textbf{7.32}}  & \textbf{25.20} / \llong{\textbf{6.81}} \\
+ \model{} w/o step 2 & 26.99 / \llong{7.49} & 36.90 / \llong{7.22} & 25.03 / \llong{6.66} \\
\midrule
Few-shot (FS) & 27.13 / \llong{7.66} & 39.47 / \llong{7.12} & 3.98 / \llong{1.34} \\
\emph{\textbf{+ \model{}}} & \textbf{30.65} / \llong{\textbf{7.72}} & \textbf{41.36} / \llong{\textbf{7.24}} & \textbf{25.05} / \llong{\textbf{6.65}} \\
+ \model{} w/o step 2 & 30.37 / \llong{7.70} & 35.54 / \llong{6.28} & 24.15 / \llong{5.82} \\
\bottomrule
\end{tabular}
}
\caption{\small{Mistral ROUGE-L / \llong{GPT-4o-Judge} main ablation study with \model{} when Step 2 is skipped.}}
\label{tab:ablation-skip-step2}
\end{table}

From \Cref{table:main-ablation-results}, we identify the unique contributions of each step within \model{}. Specifically, omitting Step~1 transforms it into OCG, whereas excluding Step~3 yields MG, and skipping Step~4 becomes MG-OCG. Here, we investigate \model{} effectiveness when skipping Step~2, Metrics' scores collection: for selected metrics from Step~1, we directly task the models to optimize them for the generated answers.  We experiment with Mistral on SAMSum, SWiPE, and CommGen-Chall. datasets because for these datasets, the best guideline combination includes MG.

The results in \Cref{tab:ablation-skip-step2} verify that without Step~2, the model performs worse, particularly for SAMSum and SWiPE in the zero-shot setting. We attribute these drops to the incorrect task properties captured and metric conflicts. A case study is provided in Appx.-\Cref{fig:skip-step-2}.

\subsection{\model{} on Real-Life Chat LLM Benchmark}

\begin{table}
\centering
\scalebox{.85}{
\begin{tabular}{lccc}
\toprule 
\textbf{Methods} & \textbf{LC Win Rate} & \textbf{Win Rate} \\
\midrule
Zero-shot (ZS) & 11.08\% & 3.17\% \\
+ OCG & 4.73\% & 2.44\% \\
+ MG & \textbf{19.13\%} & \textbf{7.07\%} \\
+ MG-OCG & 8.42\% & 3.90\% \\
{\textbf{+ \model{}}} & \textbf{19.13\%} & \textbf{7.07\%} \\
\midrule
Few-shot (FS) & 8.08\% & 2.68\% \\
+ OCG & 7.73\% & 3.45\% \\
+ MG & \textbf{12.65\%} & \textbf{4.88\%} \\
+ MG-OCG & 12.63\% & 4.88\%
 \\
{\textbf{+ \model{}}} & \textbf{12.65\%} & \textbf{4.88\%} \\
\bottomrule
\end{tabular}
}
\caption{\small{AlpacaEval2 experiments.}}
\label{tab:alpacaeval-experiments}
\vspace{-3mm}
\end{table}

We evaluate the effectiveness of \model{} in aligning LLMs with desired real-world chats. Our experiments are conducted on a subset of 203 random samples from the AlpacaEval2 benchmark  \citep{dubois2024lengthcontrolled} with ChatGPT (1106). Since AlpacaEval2 lacks training data, we select 5 random samples from the public Alpaca-GPT4 instruction-tuning dataset (\href{https://huggingface.co/datasets/vicgalle/alpaca-gpt4?row=4}{alpaca-gpt4}), despite it being relatively out-of-distribution (OOD) compared to AlpacaEval2.

\Cref{tab:alpacaeval-experiments} presents our findings. Few-shot demonstrations and OCG negatively impact performance, likely due to the OOD nature of Alpaca-GPT4 compared to AlpacaEval2. In contrast, with just $5$ Alpaca-GPT4 samples, MG metrics, and \model{} enhance performance by capturing certain response properties from GPT-4 \citep{openai2023gpt4}, nearly doubling the zero-shot points.

\section{Related Work}
\paragraph{Automatic prompt design for long-form generation.} 

Long-form generation tasks are essential and have been studied extensively \citep{li2024pre}. With LLM advancements, adapting these models for such tasks using prompt-based methods is critical yet challenging. Previous studies \citep{bang-etal-2023-multitask,yang2023exploring,hadi2023survey,zhou2023can,pan2024lost} highlight the limited efficacy of LLMs in producing outputs that resemble ground truths, as evaluated by ROUGE-L \citep{lin-2004-rouge}. Our approach autonomously composes supplementary contexts, integrating text evaluation metrics and format constraints. In addition, studies regarding enhancing instructions for LLMs \citep{wang-etal-2022-super,yin-etal-2023-read,long-etal-2024-multi-expert,wang2024enhancing}, automatic prompt optimization \citep{zhou2023large,pryzant-etal-2023-automatic}, and demonstration selection \citep{yang-etal-2023-representative,qin2023context} are also related areas that can be developed in parallel and combined with our work (\Cref{sec:longguide-can-be-combined}).

\paragraph{Controllable generation with LLMs.} 
Controllable generation during fine-tuning has been studied extensively \citep{fan-etal-2018-controllable,lakew-etal-2019-controlling,martin-etal-2020-controllable,he-etal-2022-ctrlsum}. More recently, researchers have explored prompting methods to control LLM generation. For instance, \citet{sun2023evaluating} found that LLMs struggle to meet fine-grained hard constraints, while \citet{fonseca2024can} proposed controlling stylistic features like keywords and narrative during generation, leading to improved LLM summarization outcomes. Although \citep{lu-etal-2023-bounding,fonseca2024can} are closely related to our OCG, our approach goes beyond summarization and open-ended only features, as discussed in \Cref{sec:method}. We focus on universally applicable features.

\section{Conclusion}

In this paper, we demonstrate that in-context learning (ICL) falls short in implicitly ensuring that large language models (LLMs) consistently preserve essential language and format properties in long-form generation tasks. To address this challenge, we introduce \model{}, an efficient algorithm that automatically learns the critical language and format properties from task-specific data, converting them into textual guidelines for LLMs. Our results show that \model{} significantly improves LLM performance across seven generation tasks and is highly generalizable, offering strong potential for various downstream applications with minimal data. This work paves the way for adaptive, task-specific prompt generation, advancing LLM adaptation.


\section*{Generalizability and Customization of \model{}} \label{ssec:generalizability-customization}

\model{} facilitates flexible generalization that allows customization and extension of guidelines MG and OCG for specific tasks, which we strongly recommend. For instance, in summarization, MG can focus on only 4-5 standard metrics from $S$ while integrating summary-specific metrics like ``Summary Structure'' and ``Retention of Core Supporting Evidence.'' Simultaneously, OCG can impose stricter constraints on topics, keywords, grammar, or tones \citep{fan-etal-2018-controllable,lakew-etal-2019-controlling,martin-etal-2020-controllable}. Although \model{} is primarily presented for general long-form generation, we strongly advise for these customizations to enhance its effectiveness.

\section*{Limitations} \label{sec:limitations}

Our study has several limitations. Firstly, our theoretical analysis focuses solely on the task language distribution which is $P_\mathcal{M}(X)$ or $P_\mathcal{M}(X | D_f)$ instead of the actual output distribution, which is $\arg \max_{y \in \mathcal{Y}} P_\mathcal{M}(Y = y \mid X)$ or $\arg \max_{y \in \mathcal{Y}} P_\mathcal{M}(Y = y \mid D_f, X)$. In our study, while leveraging the task language distribution allows us to hypothesize and highlight the limitations of demonstrations, shifting focus to the actual output distribution could yield more insights.

Secondly, \model{}'s learned guidelines are based on task-level and average statistics rather than sample-based details. We designed our framework at the task level to address limited data constraints, as we found that sample-based learning under these conditions leads to high errors. While task-level guidelines already demonstrate significant improvements for LLMs, sample-based guidelines could offer more tailored guidance, potentially leading to optimal results. Moreover, this average guidance approach may be ineffective for tasks with high variance in the statistics that \model{} learns. In such cases, the final step of \model{} can prevent performance decline by likely choosing no guideline. For example, we found this applies to Code2Text \citep{richardson-etal-2017-code2text} \& StoryGeneration \citep{fan-etal-2018-hierarchical}.

Thirdly, \model{} relies on models having a certain level of task knowledge to perform self-evaluation effectively, and \model{} necessitates LLMs with strong instruction-following capabilities. However, we anticipate that cutting-edge AI language models will overcome this limitation both now and in the near future. 


Lastly, the guidelines learned by \model{} may not be useful for the tasks the models are trained on. This is because these guidelines might introduce out-of-distribution context relative to the training data, thereby reducing the effectiveness of the testing inference. For instance, while we see notable enhancements on the CommonGen-Challenge dataset \citep{lin-etal-2020-commongen}, it's intriguing that we don't observe any improvements on the WebNLG \citep{gardent-etal-2017-webnlg} and E2E NLG \citep{puzikov-gurevych-2018-e2e} datasets, despite their expected similarity. Given the popularity of these datasets, we suspect the models we tested may have been previously trained on them.

\section*{Ethical Considerations}
This method could be misused to optimize prompts for harmful purposes such as generating misinformation, hate speech, or privacy violations. While our method is not intended for such uses, it is impossible to completely prevent misuse. Although our method could enhance the efficiency and efficacy of bad actors, we do not anticipate that \model{} is inherently more effective in these negative contexts than in positive applications. Finally, we employ annotators at an hourly rate of \$20, which exceeds the local minimum wage requirements.

\section*{Acknowledgement}
This research is supported by the National Research Foundation Singapore under the AI Singapore Programme (AISG Award No: AISG2-GC-2022-005, AISG Award No: AISG2-TC2023-010-SGIL) and the Singapore Ministry of Education Academic Research Fund Tier 1 (Award No: T1 251RES2207). DXL is supported by the A*STAR Computing and Information Science (ACIS) scholarship. We thank members of WING and Deep Learning Lab at NUS and the ACL RR anonymous reviewers for the constructive feedback.

\bibliography{custom,anthology}

\onecolumn
\newpage
\appendix

\section{\model{} Algorithm}

\Cref{algo:longguide} outlines the \model{} algorithm, which generates task-specific guidelines to enhance LLM performance. It selects relevant evaluation metrics, computes self-evaluated scores, formulates metric-based and structural constraints, and automatically chooses the most effective guideline. For detailed step descriptions, see \Cref{sec:method}.

\begin{algorithm*}[t]
\renewcommand{\algorithmicrequire}{\textbf{Input:}}
\renewcommand{\algorithmicensure}{\textbf{Output:}}
\caption{{\model{}}}
\label{algo:longguide}
\begin{algorithmic}[1]
\REQUIRE $\mathcal{M}$ and its generation func. $G_{\mathcal{M}}$, train data $D^{train} = \{(x^{t}_i, y^{t}_i)\}_{i=1}^{n}$, linguistic processor $L$.
\REQUIRE Task instruction $I$, instruction to select metrics $I_M$, score metrics $I_{score}$, generate MG $I_{MG}$. \\
\KwStep{1}{\textbf{Metric Collection \& Selection}} \\
\STATE Collect the set of widely-used evaluation metrics $S$\\
\STATE $M$ = [] \Comment{the set of selected metrics}\\
\FOR {$K$ training iterations}
\STATE Sample a batch $B$ from $D^{train}$
\STATE $S_{sub} := G_{\mathcal{M}}([I_M, B, S])$ \Comment{top-5 metrics selected from $S$ for best evaluating $B$} 
\STATE $M = M \cup S_{sub}$
\ENDFOR
\STATE $M = M.sort()$ \\
\KwStep{2}{\textbf{Metric Score Collection via Self-evaluation}} 
\STATE $s_{M_1} = \cdots = s_{M_m} = 0$ \Comment{the self-evaluated average scores of selected metrics}\\
\FOR {i, $(x,y)$ in enumerate($D^{train})$}
\STATE $\{s^i_{{M_1}},...,s^i_{{M_m}}\} := G_\mathcal{M}([I_{score}, x, y, M])$ \Comment{self-evaluation}\\
\STATE Update $s_{M_j} = s_{M_j} + (s^i_{{M_j}} - s_{M_j})/(i + 1)$ for all j in range(m)
\ENDFOR
\STATE $score_{M}=[s_{M_1},\cdots,s_{M_m}]$ \\
\KwStep{3}{\textbf{Generating Metric Guideline}}
\STATE $\{d_{M_1},...,d_{M_m}\} := G_\mathcal{M}([I_{MG}, scores_{M}, M])$ \Comment{generate metrics' definitions w.r.t scores}\\
\STATE \textcolor{blue}{MG = joined with newline ($\{d_{M_1},...,d_{M_m}\}$)} \\
\KwStep{4}{\textbf{Output Constraint Guideline}}
\STATE Using $L$ to compute $(min_s, max_s, avg_s)$ of \#sentences in $y^t_i$, and $(min_t, max_t, avg_t)$ of \#tokens\\
\STATE \textcolor{blue}{OCG = ``The {response} must have from \{$min_s$\} to \{$max_s$\} sentences and from \{$min_t$\} to \{$max_t$\} words with an average of \{$avg_t$\} words and \{$avg_s$\} sentences.''}\\
\KwStep{5}{\textbf{MG–OCG selection}}
\STATE $G = \{$w/o guideline, $MG, OCG, MG$ \& $OCG\}$\\
\STATE \textcolor{blue}{$\model{}=\arg\max_{g \in G}(performance(\mathcal{M} | I, g, D^{train}))$}\Comment{automatic guideline selection}\\
\ENSURE \model{}
\end{algorithmic}
\end{algorithm*}

\section{Theoretical Intuitions} \label{ssec:theoretical-analysis}


We now present a theoretical analysis to explain the observed phenomena.

\paragraph{Task formulation.} A long-form generation train dataset for a task $T$ with $n$ samples is defined as $D = \{(x^t_i,y^t_i)\}_{i=1}^{n}$, where $x^t_i$ and $y^t_i$ are the input context and ground-truth {sentence- or paragraph-long} responses. In the \textbf{instruction-based} setting, an LLM $\mathcal{M}$ is expected to generate $y$ given $x$ and an input {i}nstruction $I$. Recall that $P_{\mathcal{M}}$ and $P_{T}$ are the probability functions of $\mathcal{M}$'s generation and the task data distribution, then:

\begin{remark} \label{theorem:theorem}
Under mild assumptions and $P_{\mathcal{M}} \neq P_T$, there exists $x \in \mathcal{X}$ such that $P_{\mathcal{M}}(X = x|D_f) \neq P_T(X=x)$.
\end{remark} 

The proof is provided in \Cref{appx:proof-theorem1}. In essence, \Cref{theorem:theorem} asserts that when $\mathcal{M}$ fails to capture the true task data distribution, $d_f$ cannot recover the desired alignment in the limit. As a result, certain task language and format properties, even when well-presented in demonstrations, may not be implicitly captured and preserved during LLM generation. We term this unsolved issue the \textbf{text property transfer (PT)} challenge: ensuring that $\mathcal{M}$ captures and preserves specific desired text properties observed in a limited set of labeled data to responses.  
\emph{We hypothesize that explicitly guiding the model with essential text property criteria mitigates the mismatch identified in \Cref{theorem:theorem} between the induced and task-specific language distributions, leading to improved performance.} 

\begin{definition}{\textbf{}} \label{defi:defi}
For the task $T \triangleq \{D, \mathcal{L}\}$, a text property  $T_1$ of task $T$ is defined as the task $T_1 \triangleq \{D_1, \mathcal{L}_1\}$, where $D_1=\{(x^t_i, f_1(y^t_i))\}_{i=1}^n$
is the train data of $T_1$ mapped from $D$ by a text property metric (\emph{feature map}) $f_1: \mathcal{Y} \to \mathbb{R}$, with $\mathcal{L}$ and $\mathcal{L}_1$ being the objectives of $T$ and $T_1$.
\end{definition}

Suppose that the objective is $\textit{minimize}_{\theta\in \Theta}{\mathcal{L}(\theta, T)}$ with $\theta$ being a tunable factor of $\mathcal{M}$, which can be its parameters or input instruction, and $\Theta$ being its space. We propose the following hypothesis verified in \Cref{ssec:main-findings}:

\begin{hypothesis}\label{hypothesis:hypothesis2}
$T$ can be approximated by $r$ well-chosen text property tasks $T_1, ..., T_r$ ($T \approx T_1\oplus...\oplus T_r$) with corresponding objectives $\mathcal{L}_1,..,\mathcal{L}_r$ 
such that by jointly optimizing them during the generation process, we can approximately optimize $T$'s objective, i.e., $\arg \min_{\theta \in \Theta} \mathcal{L} \approx \arg \min_{\theta \in \Theta} \sum_{i=1}^{r} \mathcal{L}_i$.
\end{hypothesis}

\subsection{Proof of \Cref{theorem:theorem}} \label{appx:proof-theorem1}

\begin{assum} \label{assum:assum}
There exists $x \in \mathcal{X}$ for which $P_\mathcal{M}(X=x) \neq P_{T}(X=x)$.
\end{assum}

This assumption is intuitive and realistic, recognizing that LLMs cannot fully capture the vast and nuanced complexity of real-world language beyond their training data. It contradicts the common assumption  $P_\mathcal{M}(X) = P_T(X)$ made by prior studies \citep{xie2021explanation,wang2024large}. A simple empirical evidence is provided in \Cref{appx:case-study-assum}. We also assume:

\begin{assum} \label{assum:assum2}
Two probability functions are  {functionally zero equivalent} if they act on the same input space and any arbitrary event causes both functions to be zero or non-zero. We assume that $P_T$ and $P_\mathcal{M}$ are functionally zero equivalent, i.e., $P_\mathcal{M}(X=x) = 0 \Leftrightarrow P_T(X=x) = 0$ $\forall x \in \mathcal{X}$.
\end{assum}

Note that \Cref{assum:assum2} is a relaxed version of the common assumption $P_\mathcal{M}(X) = P_T(X)$, and does not conflict with \Cref{assum:assum}. 

\begin{proof}[Proof of \Cref{theorem:theorem}]

We prove this theorem by contradiction. Suppose the negation of \Cref{theorem:theorem} is true, i.e., there exists a $D_1 \in \mathcal{D}$ such that $\forall X \in \mathcal{X}$, $P_\mathcal{M}(X|D_1) = P_T(X)$ (\emph{S1}). 

Now, let us consider the event $X \cap D_1^{c}$ where $D_1^c$ is the conjugate of event $D_1$, or $D_1^c = \mathcal{D}\backslash{D_1}$. We have $P_\mathcal{M}(X \cap D_1^{c} | D_1) = 0$. From the assumption \emph{(S1)}, we derive $P_T(X \cap D_1^{c}) = 0$. From the \Cref{assum:assum2}, since $P_\mathcal{M}$ and $P_{T}$ are functionally zero equivalent, we have $P_\mathcal{M}(X \cap D_1^{c}) = 0$. 

Similarly, we can consider the event $X^c \cap D^c$ where $X^c$ is the conjugate of $X$, we arrive at $P_\mathcal{M}(X^c \cap D_1^c) = 0$. 

Since the two $X \cap D_1^{c}$ and $X^c \cap D_1^{c}$ form a disjoint union of $D_1^c$, we derive $P_\mathcal{M}(D_1^c) = P_\mathcal{M}(X \cap D_1^{c}) + P_\mathcal{M}(X^c \cap D_1^{c}) = 0 + 0 = 0$. Since $D_1$ and $D_1^c$ form a disjoint union of $\mathcal{D}$, we have $P_\mathcal{M}(D_1) = 1$.

From the negation statement \emph{(S1)}, we have $P_\mathcal{M}(X|D_1) = P_T(X)$ $\forall X \in \mathcal{X}$. Since $X \cap D_1$ and $X \cap D_1^c$ form a disjoint union of $X$, we have $P_\mathcal{M}(X) = P_\mathcal{M}(X \cap D_1) + P_\mathcal{M}(X \cap D_1^c) = P_\mathcal{M}(X \cap D_1) + 0 = P_\mathcal{M}(X \cap D_1)$. We also have $P_\mathcal{M}(X|D_1) = \frac{P_\mathcal{M}(X \cap D_1)}{P_\mathcal{M}(D_1)}$ from Bayes's theorem, meaning that $P_\mathcal{M}(X|D_1) = P_\mathcal{M}(X \cap D_1) = P_\mathcal{M}(X)$ (since $P_\mathcal{M}(D_1) = 1$). Meanwhile, from the negation statement \emph{(S1)}, we have $P_\mathcal{M}(X|D_1) = P_T(X)$, thus $P_\mathcal{M}(X) = P_T(X)$ for all $X \in \mathcal{X}$, which contradicts to our \Cref{assum:assum}. Therefore, our negation statement \emph{(S1)} is false, leading to \Cref{theorem:theorem} is true.
\end{proof}

\section{Empirical Case Studies Supporting \Cref{sec:demos-anayn}} \label{appx:case-studies}

\subsection{Empirical Case Study for \Cref{assum:assum}} \label{appx:case-study-assum}

Here, we provide empirical evidence for \Cref{assum:assum}. It posits that for a given long-form generation test task $T$, the language model $\mathcal{M}$ does not entirely encapsulate $T$’s true language distribution. This assumption aligns with real-world scenarios since LLMs are not infinitely powerful: there exist task instances where they are out-of-distribution (OOD) for $\mathcal{M}$.

For example, let us consider \Cref{theorem:theorem} as the task $T$ and $\mathcal{M}$ be ChatGPT \cite{openai2022chatgpt}. The true language distribution is the specific vocabulary, technical jargon, and nuanced writing style typically found in mathematics. The full dialogue can be found at \url{https://chatgpt.com/share/0189761f-202b-46e6-bab0-39a80bd5340e}.

\begin{tcolorbox}[colback=white,boxrule=0.5pt,fonttitle=\small, fontupper=\small]
\texttt{Input: Given two probability functions $P_1$ and $P_2$. Suppose that there exists at least X such that $P_1(X) \neq P_2(X)$. Prove that for any D, there exists at least X' such that $P_1(X'|D) \neq P_2(X')$.}
\end{tcolorbox}

\begin{tcolorbox}[colback=white,boxrule=0.5pt,fonttitle=\small, fontupper=\small]
\texttt{ChatGPT output (30 Jul 24): To prove the statement, we need to understand the conditions under which the equality of the probability functions \(P_1\) and \(P_2\) given a condition \(D\) breaks down.}

\texttt{Let's start by stating the given conditions more clearly:}

\texttt{1. We have two probability functions \(P_1\) and \(P_2\).}
\texttt{2. There exists at least one event \(X\) such that \(P_1(X) \neq P_2(X)\).}

\texttt{We need to show that for any event \(D\), there exists at least one event \(X'\) such that \(P_1(X'|D) \neq P_2(X'|D)\).}

\texttt{Let's break it down into steps…}
\end{tcolorbox}

Based on the above example, we can see that ChatGPT does not focus on proving 
$P_1(X'|D) \neq P_2(X')$. Instead, it misunderstood as proving \(P_1(X'|D) \neq P_2(X'|D)\), showing that the ChatGPT’s language distribution when solving the task is different from the true task language distribution.

\subsection{Empirical Case Study for \Cref{theorem:theorem}} \label{appx:case-study-theorem1}

Empirical evidence supporting \Cref{theorem:theorem} is not possible to establish since it is an extremal result concerning existence and non-existence. Essentially, \Cref{theorem:theorem} says that if at the beginning, the two distributions of the task and language model are not the same (``first not the same'') but functionally zero equivalent, then for any demonstrations, the two distributions of the task and language model conditioned on those demonstrations are not the same (``second not the same'').


It's important to note that the data point causing the ``first not the same" can differ from the data point causing the ``second not the sam'', and this ``second not the same'' data point needs to be examined by all possible demonstrations. This makes it difficult to empirically verify the theorem since the demonstration space is vast.

\section{\model{}'s Extra Preliminary Properties} \label{Appx:longguide-properties}

\subsection{\model{} can Improve Non-instruct Models} \label{sec:can-longguide-improve-non-instruct}

Using guidelines learned by \model{}, we add more instructions to models. Therefore, we aim to examine whether non-instruct models can benefit from these guidelines. Our final conclusion is yes, \model{} has strong potential to enhance non-instruct models. 

\begin{wraptable}{r}{0.6\textwidth}
\vspace{-3mm}
\footnotesize
\scalebox{0.9}{
\begin{tabular}{lccc}
\toprule 
\textbf{Methods} & \textbf{CNN (3.0.0)}  & \textbf{IWSLT17} & \textbf{CommGen-Chall.}\\
\midrule
Zero-shot (ZS) & 7.60$_{\pm0.58}$ & 2.99$_{\pm0.83}$ & \textbf{10.96}$_{\pm0.36}$ \\
\hdashline
+ OCG & 6.60$_{\pm0.74}$\textcolor[RGB]{250,0,0}{$\downarrow$} & 3.70$_{\pm0.29}$\textcolor[RGB]{0,80,71}{$\uparrow$} & 10.12$_{\pm0.56}$\textcolor[RGB]{250,0,0}{$\downarrow$} \\
+ MG & \textbf{9.04}$_{\pm1.02}$\textcolor[RGB]{0,80,71}{$\uparrow$} & \textbf{5.39}$_{\pm0.93}$\textcolor[RGB]{0,80,71}{$\uparrow$} & 8.55$_{\pm0.74}$\textcolor[RGB]{250,0,0}{$\downarrow$}\\
+ MG-OCG & 8.38$_{\pm0.91}$\textcolor[RGB]{0,80,71}{$\uparrow$} & 4.59$_{\pm0.97}$\textcolor[RGB]{0,80,71}{$\uparrow$} & 7.99$_{\pm0.70}$\textcolor[RGB]{250,0,0}{$\downarrow$} \\
\hdashline
+ \model{} & \textbf{9.04}$_{\pm1.02}$\textcolor[RGB]{0,80,71}{$\uparrow$} & \textbf{5.39}$_{\pm0.93}$\textcolor[RGB]{0,80,71}{$\uparrow$} & \textbf{10.96}$_{\pm0.36}$\\
\midrule
Few-shot (FS) & 3.14$_{\pm0.32}$ & 3.44$_{\pm0.83}$ & 4.67$_{\pm0.33}$ \\
\hdashline
+ OCG & 2.24$_{\pm0.21}$\textcolor[RGB]{250,0,0}{$\downarrow$} & 3.86$_{\pm0.61}$\textcolor[RGB]{0,80,71}{$\uparrow$} & 8.11$_{\pm0.63}$\textcolor[RGB]{0,80,71}{$\uparrow$} \\
+ MG & \textbf{3.24}$_{\pm0.26}$\textcolor[RGB]{0,80,71}{$\uparrow$} & 6.65$_{\pm0.97}$\textcolor[RGB]{0,80,71}{$\uparrow$} & \textbf{10.71}$_{\pm0.80}$\textcolor[RGB]{0,80,71}{$\uparrow$} \\
+ MG-OCG & 2.99$_{\pm0.29}$\textcolor[RGB]{250,0,0}{$\downarrow$} & \textbf{7.88}$_{\pm0.91}$\textcolor[RGB]{0,80,71}{$\uparrow$} & 9.39$_{\pm0.89}$\textcolor[RGB]{0,80,71}{$\uparrow$} \\
\hdashline
+ \model{} & 2.24$_{\pm0.21}$\textcolor[RGB]{250,0,0}{$\downarrow$} & \textbf{7.88}$_{\pm0.91}$\textcolor[RGB]{0,80,71}{$\uparrow$} & \textbf{10.71}$_{\pm0.80}$\textcolor[RGB]{0,80,71}{$\uparrow$} \\
\bottomrule
\end{tabular}
}
\caption{\small{ROUGE-L performance of \textbf{Mistral-7B-v0.1} using \model{} learned by \textbf{Mistral-7B-Instruct-v0.2}. We observe that \model{} improves more than half of the experiments, showing its potential effectiveness in enhancing even non-instruct models, especially for the translation task.}}
\vspace{-5mm}
\label{tab:longpo-improve-non-instruct}
\end{wraptable}

\paragraph{Setups.} Since non-instruct models might struggle to follow our instructions to generate the guidelines \Cref{sec:limitations}, we utilize the guidelines learned by an instruct model instead. We run our experiments with \textbf{Mistral-7B-v0.1}\footnote{\url{https://huggingface.co/mistralai/Mistral-7B-v0.1}}\citep{jiang2023mistral} using the guidelines learned by Mistral-7B-Instruct-v0.2. 

\paragraph{Findings.} The results are provided in \Cref{tab:longpo-improve-non-instruct}. We observe
that LongGuide improves more than half of the experiments, showing its potential effectiveness in enhancing
even non-instruct models, especially for the translation task.

\subsection{\model{} can be Transferable from Weaker to Stronger Models} \label{appdx:longguide-is-transferrable}

\begin{wraptable}{r}{0.6\textwidth}
\vspace{-3mm}
\scalebox{.65}{
\begin{tabular}{lccc}
\toprule 
\textbf{Methods} & \textbf{CNN (3.0.0)} & \textbf{IWSLT17 en-ja} & \textbf{CommGen-Chall.}\\
\midrule
ChatGPT Zero-shot (ZS) & 20.12$_{\pm 0.27}$ & 36.13$_{\pm 0.87}$ & 24.21$_{\pm 0.37}$ \\
ChatGPT ZS w/ Mistral's MG & 21.41$_{\pm 0.62}$\textcolor[RGB]{0,80,71}{$\uparrow$} & 39.66$_{\pm 2.47}$\textcolor[RGB]{0,80,71}{$\uparrow$} & 29.95$_{\pm 23.66}$\textcolor[RGB]{0,80,71}{$\uparrow$} \\
\midrule
ChatGPT Few-shot (FS) & 14.51$_{\pm 0.80}$ & 31.93$_{\pm 1.88}$ & 22.08$_{\pm 0.63}$ \\
ChatGPT FS w/ Mistral's MG & 13.96$_{\pm 11.50}$\textcolor[RGB]{250,0,0}{$\downarrow$} & 32.34$_{\pm 13.79}$\textcolor[RGB]{0,80,71}{$\uparrow$} & 33.34$_{\pm 13.56}$\textcolor[RGB]{0,80,71}{$\uparrow$} \\
\midrule
\midrule
Mistral Zero-shot (ZS) & 19.23$_{\pm 0.34}$ & 13.12$_{\pm 1.39}$ & 10.12$_{\pm 0.02}$ \\
Mistral w/ ChatGPT's MG  & 19.67$_{\pm 0.71}$\textcolor[RGB]{0,80,71}{$\uparrow$} & 7.98$_{\pm 1.49}$\textcolor[RGB]{250,0,0}{$\downarrow$} & 6.29$_{\pm 1.06}$\textcolor[RGB]{250,0,0}{$\downarrow$} \\
\midrule
Mistral Few-shot (FS) & 17.56$_{\pm 0.63}$ & 12.69$_{\pm 1.82}$ & 3.89$_{\pm 0.17}$\\
Mistral FS w/ ChatGPT's MG & 19.00$_{\pm 7.82}$\textcolor[RGB]{0,80,71}{$\uparrow$} & 11.86$_{\pm 2.79}$\textcolor[RGB]{250,0,0}{$\downarrow$} & 3.61$_{\pm 0.38}$\textcolor[RGB]{250,0,0}{$\downarrow$}\\
\bottomrule
\end{tabular}
}
\caption{\small{\model{} can be transferable from weaker to stronger models, evaluated by ROUGE-L.}}
\vspace{-3mm}
\label{tab:longpo-transferable-guidelines}
\end{wraptable}

We find that the guidelines learned by \model{} are transferable from weaker to stronger models. A weaker model can learn the guidelines at a low cost, which can then be used to enhance the performance of stronger models. This is particularly advantageous because powerful models are often closed-source and expensive to query, whereas open-source models are weaker but free to use. 

\paragraph{Setups.} We demonstrate this through experiments on CNN (3.0.0), IWSLT17 en-ja, and CommGen-Chall, representing all the tasks. We used the MG generated by Mistral for experiments on ChatGPT and vice versa under both zero-shot and few-shot settings. 

\paragraph{Findings.} \Cref{tab:longpo-transferable-guidelines} outlines the results. We observe that Mistral's MG generally improves ChatGPT performance, but not vice versa. Explaining these phenomena, firstly, the OCG is transferable across models because it is independent of any specific model. Secondly, the MG, while it helps models capture task distributions, an MG learned from a stronger model may not benefit a weaker model, as the weaker model may misinterpret it. In contrast, the stronger model, with better text comprehension, can generalize task distributions from MG even when MG is poor and/or not well expressive generated by the weaker model.

\subsection{\model{} can be Compared and Combined with Automatic Prompt Optimization Algorithms} \label{sec:longguide-can-be-combined}

The MG and OCG learned by \model{} may not be fully optimized for LLMs. Hence, it's intuitive to suggest that LLMs could achieve even greater performance by adopting optimal guidelines. In this section, we illustrate that the guidelines learned by \model{} can be further refined through discrete prompt optimization algorithms. This capability is advantageous for \model{}, enabling its concurrent development and integration with automatic prompt optimization algorithms.

\paragraph{Setup.} We employ two strong prompt optimizers, APO \citep{pryzant-etal-2023-automatic} and adv-ICL \citep{do2024prompt}, in our experiments. \llong{We also compare \model{} with EvolPrompt \citep{guo2024connecting} in this section.} Here is our methodology: we integrated the guidelines generated by \model{} into the prompt, including the input instruction and demonstrations. Subsequently, we applied the prompt optimizers to refine the input instruction, demonstrations, and guidelines. Our experiments were conducted using Mistral on datasets including CNN, IWSLT 2017 en-ja, and CommonGen-Challenge. Following our findings in \Cref{table:main-ablation-results}, the guideline being optimized for CNN and IWSLT 2017 en-ja is OCG, while for CommonGen-Challenge is MG-OCG.

\begin{wraptable}{r}{0.6\textwidth}
\footnotesize
\vspace{-3mm}
\scalebox{0.8}{
\begin{tabular}{lccc}
\toprule 
\textbf{Methods} & \textbf{CNN (3.0.0)}  & \textbf{IWSLT17} & \textbf{CommGen-Chall.}\\
\midrule
Zero-shot (ZS) & 19.23$_{\pm 0.34}$ & 13.12$_{\pm 1.39}$ & 10.12$_{\pm 0.02}$ \\
+ APO & 19.53$\pm_{2.08}$ & 14.45$\pm_{1.84}$ & 11.21$\pm_{2.02}$ \\
\llong{+ EvolPrompt} & \llong{20.16$\pm_{3.44}$}  & \llong{15.04$\pm_{2.12}$} & \llong{14.06$\pm_{3.02}$} \\
+ adv-ICL & 18.87$\pm_{2.69}$ & 15.01$\pm_{1.72}$ & 13.12$\pm_{2.21}$ \\
\hdashline
+ \model{} & 22.46$_{\pm 0.64}$ & 16.53$_{\pm 0.59}$ & 25.20$_{\pm 1.89}$ \\
\midrule 
+ \model{} + APO & \textbf{22.76}$_{\pm 1.04}$\textcolor[RGB]{0,80,71}{$\uparrow$} & \textbf{17.13}$_{\pm 1.05}$\textcolor[RGB]{0,80,71}{$\uparrow$} & \textbf{27.01}$_{\pm 1.01}$\textcolor[RGB]{0,80,71}{$\uparrow$} \\
+ \model{} + adv-ICL & 21.97$_{\pm 3.21}$\textcolor[RGB]{250,0,0}{$\downarrow$} & 16.90$_{\pm 2.15}$\textcolor[RGB]{0,80,71}{$\uparrow$} & 26.18$_{\pm 3.47}$\textcolor[RGB]{0,80,71}{$\uparrow$} \\
\bottomrule
\end{tabular}
}
\caption{\small{Guidelines learned by \model{} are further optimized by discrete prompt optimization frameworks bringing even better performance, with Mistral, evaluated by ROUGE-L.}}
\vspace{-3mm}
\label{tab:longpo-with-po-algorithms}
\end{wraptable}

\paragraph{Findings.} 
Our results are detailed in \Cref{tab:longpo-with-po-algorithms}. In summary, when further optimizing the OCG using APO and adv-ICL for CNN and IWSLT 2017, we observed a slight improvement. This could be attributed to the OCG already being concise and straightforward, making it easier for models to grasp. However, for the CommonGen-Challenge dataset, which utilizes the MG-OCG guideline with more detail, APO and adv-ICL have a greater amount of material to optimize within the prompts. This led to a substantial improvement in performance compared to the other datasets.

\section{Supplementary Results and Discussions}
\label{Appx:extra-analysis}

\begin{table}
\centering
\footnotesize
\scalebox{.8}{
\begin{tabular}{cl|ccc|c|c|c|c|c}
\toprule
 & &  & \textbf{Sum.} & & \textbf{Simplification} & \textbf{Translation} & \textbf{Dialogue Gen.}  &\textbf{Table2Text} & \\
\midrule
& \textbf{Method} & \textbf{SAMSum} & \textbf{CNN (3.0.0)} & \textbf{XL-Sum} & \textbf{SWiPE} & \textbf{IWSLT17 en-ja} & \textbf{Syn. Persona} & \textbf{Comm.-Chall.} & \emph{\textbf{Avg.}} \\
\midrule
& \textbf{\#shots (ran.)} &  3 & 3 & 5 & 3 & 3 &  5 & 5 & \\
\midrule
\multirow{9}*{\begin{tabular}[c]{@{}l@{}}{\rotatebox{90}{\textbf{Mistral-it (0.2)}}}
\end{tabular}} 
& Zero-shot (ZS) &  22.20 / 20.05  & 19.23 / 20.43 & 9.19 / 8.82 &  36.60 / 39.01 &  13.12 / 13.72 &  12.76 / 11.79 &  10.12 / 6.19 & \emph{17.38} \\
&+ APO & 23.77 / 22.02 & 19.53 / 21.46 & 12.06 / 11.50 & 36.92 / 39.41 & 14.45 / 15.49 & 10.66 / 10.05 & 11.21 / 7.12  & \emph{18.26} \\
\cdashline{2-10}
&\textbf{\emph{+ \model{}}} & \textbf{28.35} / \textbf{28.79} & \textbf{22.46} / \textbf{27.82} & \textbf{14.38} / \textbf{14.13} & \textbf{38.21} / \textbf{40.83} & \textbf{16.53} / \textbf{18.81} & \textbf{14.69} / \textbf{12.86} & \textbf{25.20} / \textbf{24.03} & \emph{\textbf{23.37}}  \\
& \textcolor{gray}{\emph{\% gain (+)}} & \textcolor[RGB]{0,80,71}{$6.15$} / \textcolor[RGB]{0,80,71}{$8.74$} & \textcolor[RGB]{0,80,71}{$3.23$} / \textcolor[RGB]{0,80,71}{$7.39$}  & \textcolor[RGB]{0,80,71}{$5.19$} / \textcolor[RGB]{0,80,71}{$5.31$} & \textcolor[RGB]{0,80,71}{$1.61$} / \textcolor[RGB]{0,80,71}{$1.82$} & \textcolor[RGB]{0,80,71}{$3.41$} / \textcolor[RGB]{0,80,71}{$5.09$} & \textcolor[RGB]{0,80,71}{$1.93$} / \textcolor[RGB]{0,80,71}{$1.07$} & \textcolor[RGB]{0,80,71}{$15.08$} / \textcolor[RGB]{0,80,71}{$17.84$} & \emph{\textcolor[RGB]{0,80,71}{5.99}} \\
\cmidrule{2-10}
&Few-shot (FS)  & 27.13 / 27.21 & 17.56 / 20.55 & 9.79 / 8.32 & 39.47 / 39.76 & 12.69 / 13.78 &  3.56 / 2.67 & 3.98 / 1.94 & \emph{16.32} \\
&+ APO & 26.23 / 25.88 & 18.18 / 21.32 & 11.99 / 11.71 & 39.55 / 39.56 & 14.08 / 14.70 & 4.26 / 2.91 & 5.45 / 3.76 & \emph{17.12} \\
\cdashline{2-10}
& \textbf{\emph{+ \model{}}} & \textbf{30.65} / \textbf{31.72} & \textbf{19.19} / \textbf{22.30} & \textbf{15.23} / \textbf{14.02} & \textbf{41.36} / \textbf{41.22} & \textbf{16.62} / \textbf{17.92} & \textbf{5.25} / \textbf{4.46} & \textbf{25.05} / \textbf{21.90} & \emph{\textbf{21.92}} \\
& \textcolor{gray}{\emph{\% gain (+)}} & \textcolor[RGB]{0,80,71}{$3.52$} / \textcolor[RGB]{0,80,71}{$4.51$} & \textcolor[RGB]{0,80,71}{$1.63$} / \textcolor[RGB]{0,80,71}{$1.75$} & \textcolor[RGB]{0,80,71}{$5.44$} / \textcolor[RGB]{0,80,71}{$5.70$}  & \textcolor[RGB]{0,80,71}{$1.89$} / \textcolor[RGB]{0,80,71}{$1.46$} & \textcolor[RGB]{0,80,71}{$3.66$} / \textcolor[RGB]{0,80,71}{$4.14$} & \textcolor[RGB]{0,80,71}{$1.69$} / \textcolor[RGB]{0,80,71}{$1.79$} & \textcolor[RGB]{0,80,71}{$21.07$} / \textcolor[RGB]{0,80,71}{$19.96$} & \emph{\textcolor[RGB]{0,80,71}{5.61}} \\
\midrule
\multirow{7}*{\begin{tabular}[c]{@{}l@{}} {\rotatebox{90}{\textbf{ChatGPT}}} \\
\end{tabular}} 
& Zero-shot (ZS) &  23.83 / 20.23  & 20.12 / 24.11 & 10.80 / 11.46 &  45.09 / 43.28 &  36.13 / 38.32 &  19.46 / 19.75 & 24.21 / 24.04 & \emph{25.77} \\
&+ APO & 25.05 / 22.90 & 20.34 / 21.88 & 12.19 / 12.52 & \textbf{46.32} / \textbf{44.89} & 37.74 / 39.01 & 19.91 / 19.80 & 23.63 / 24.18 & \emph{26.45} \\
\cdashline{2-10}
& \textbf{\emph{+ \model{}}} & \textbf{30.47} / \textbf{28.37} & \textbf{22.19} / \textbf{30.79} & \textbf{20.93} / \textbf{22.61} & 45.09 / 43.28 & \textbf{41.22} / \textbf{43.79} & \textbf{22.98} / \textbf{23.79} & \textbf{34.41} / \textbf{36.84} & \emph{\textbf{31.91}} \\
& \textcolor{gray}{\emph{\% gain (+)}} & \textcolor[RGB]{0,80,71}{$6.64$} / \textcolor[RGB]{0,80,71}{$8.14$}  & \textcolor[RGB]{0,80,71}{$2.07$} / \textcolor[RGB]{0,80,71}{$6.68$} & \textcolor[RGB]{0,80,71}{$10.13$} / \textcolor[RGB]{0,80,71}{$11.15$} & \textcolor[RGB]{0,80,71}{$0.00$} / \textcolor[RGB]{0,80,71}{$0.00$} & \textcolor[RGB]{0,80,71}{$5.09$} / \textcolor[RGB]{0,80,71}{$5.47$} & \textcolor[RGB]{0,80,71}{$3.52$} / \textcolor[RGB]{0,80,71}{$4.04$} & \textcolor[RGB]{0,80,71}{$10.20$} / \textcolor[RGB]{0,80,71}{$12.80$} & \emph{\textcolor[RGB]{0,80,71}{6.13}} \\
\cmidrule{2-10}
&Few-shot (FS) & 22.21 / 25.37  & 14.51 / 17.52 &  11.42 / 10.83 & 33.72 / 32.69 & 31.93 / 32.68 & 16.10 / 18.10 & 22.08 / 23.52  & \emph{22.34} \\
&+ APO & 24.22 / 22.77 & 15.20 / 17.04 & 14.07 / 15.69 & 34.46 / 33.18 & 33.72 / 35.50 & 17.68 / 17.77 & 25.09 / 24.70 & \emph{23.65} \\
\cdashline{2-10}
& \textbf{\emph{+ \model{}}} & \textbf{31.46} / \textbf{30.04}  & \textbf{18.17} / \textbf{18.52} & \textbf{19.95} / \textbf{22.49}  & \textbf{37.60} / \textbf{35.66} & \textbf{38.43} / \textbf{42.84} & \textbf{22.36} / \textbf{20.31} & \textbf{38.21} / \textbf{37.64} & \emph{\textbf{29.55}} \\
& \textcolor{gray}{\emph{\% gain (+)}} & \textcolor[RGB]{0,80,71}{$9.25$} / \textcolor[RGB]{0,80,71}{$4.67$} & \textcolor[RGB]{0,80,71}{$3.66$} / \textcolor[RGB]{0,80,71}{$1.00$} & \textcolor[RGB]{0,80,71}{$8.53$} / \textcolor[RGB]{0,80,71}{$11.66$} & \textcolor[RGB]{0,80,71}{$3.88$} / \textcolor[RGB]{0,80,71}{$2.97$} & \textcolor[RGB]{0,80,71}{$6.50$} / \textcolor[RGB]{0,80,71}{$10.16$} & \textcolor[RGB]{0,80,71}{$6.53$} / \textcolor[RGB]{0,80,71}{$2.21$} & \textcolor[RGB]{0,80,71}{$16.13$} / \textcolor[RGB]{0,80,71}{$14.12$} & \emph{\textcolor[RGB]{0,80,71}{7.21}} \\
\bottomrule
\end{tabular}}
\caption{
\small{Supplemetary {ROUGE-L / BLEU-1} results on seven long-form generation tasks showing that the trends of ROUGE-L and BLEU-1 scores are nearly identical.}}
\label{table:supplement-main-experimental-results}
\end{table}

\subsection{Additional Baselines: Using more Shots for ICL} \label{ssec:extra-baselines-more-shots}

\begin{wraptable}{r}
{0.6\textwidth}
\footnotesize
\vspace{-5mm}
\resizebox{.6\textwidth}{!}{%
\begin{tabular}{lccc}
\toprule
\#shot & CNN (3.0.0) & SWiPE & Comm.-Chall. \\
\midrule
3-5 shots & 14.51 / 4.38 & 33.72 / 5.07 & 22.08 / 4.19 \\
+ LongGuide & \textbf{18.17 / 4.42} & \textbf{37.60 / 5.25} & \textbf{38.21 / 7.21} \\
\midrule
10-50 shots & 20.55 / 6.67 & 44.04 / 6.07 & 28.18 / 4.85 \\
+ LongGuide & \textbf{21.69 / 6.82} & \textbf{46.17 / 6.67} & \textbf{42.55 / 7.72} \\
\bottomrule
\end{tabular}%
}
\caption{Performance comparison of models with and without LongGuide across different datasets and shot settings.}
\vspace{-3mm}
\label{tab:longguide_comparison_more_shots}
\end{wraptable}

\llong{We supplement the results for CNN (3.0.0), SWiPE, and Comm.-Chall. in \Cref{tab:longguide_comparison_more_shots} where we use 10 shots for CNN, 50 shots for SWiPE, and Comm.-Chall up to the window size limit of gpt-3.5-turbo-1106 evaluated by ROUGE-L / GPT-4o-Judge scores.}

\llong{We observe that while supplementing more shots to ChatGPT improves model’s performance, LongGuide further boosts the ICL performance significantly for all three benchmarks.}

\subsection{How Does LLM Handle \model{}, and Context Given \model{}?} \label{ssec:attention-analysis}

\begin{wrapfigure}{r}{0.35\textwidth}
\small 
\centering
\vspace{-5mm}
\includegraphics[width=1\linewidth, trim={0cm 0cm 0cm 0cm},clip]{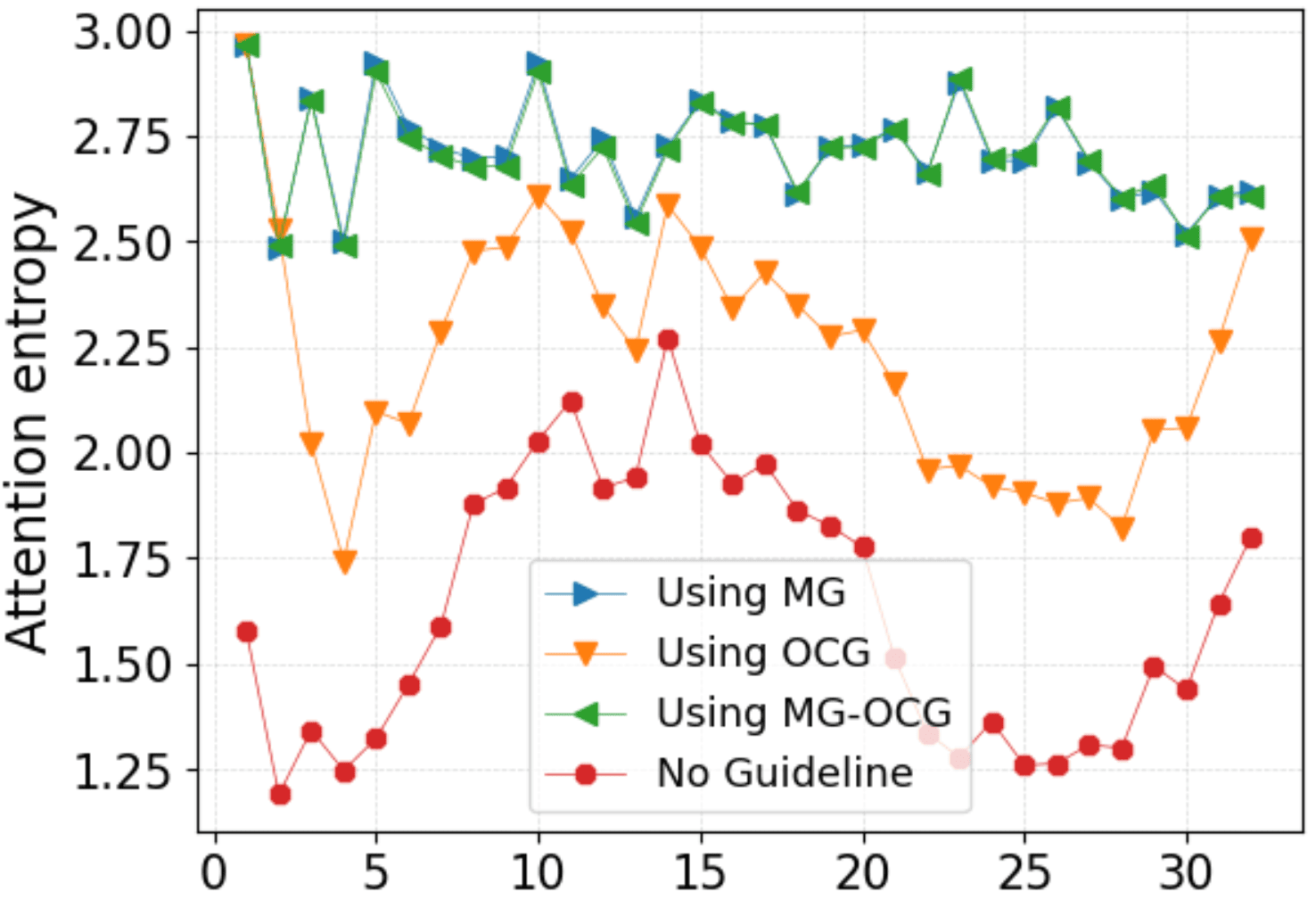}
\vspace{-3mm}
\caption{\small{Entropy of  attention over the input context across 32 Mistral layers.}}
\label{fig:entropy-attention-analysis}
\end{wrapfigure}

To analyze \model{}'s impact on LLMs, we perform a simple attention analysis to investigate (1) how LLMs attend to \model{} and (2) utilize the input context when conditioning on \model{}. Specifically, for (1), we calculate the average attention scores across all heads and layers for each guideline token. For (2), we evaluate the entropy of the attention scores overall context tokens.
We experiment with Mistral on 100 SAMSum random samples. We learn two key findings.

Firstly, Mistral shows substantial attention to the guidelines. By using MG, 37.81\% of attention is on guideline tokens. For OCG, it is 22.56\%, and MG-OCG, 37.87\%. Notably, the average attention on OCG tokens is higher than on context, while MG and MG-OCG receive a fair amount, confirming mode attention on guidelines (Appx.-\Cref{tab:attention-score-analysis}).

Secondly, from \Cref{fig:entropy-attention-analysis}, Mistral exhibits more selective context attention when conditioned on guidelines. The largest entropy gap occurs in the first layer, where with guidelines, the model sparsely processes the context but, without them, is biased towards focusing narrowly on specific context parts. In the final layer, the model distributes attention more evenly with guidelines than without. Generally, MG stabilizes context use across layers, while OCG shows greater variance, likely because it does not directly control generation quality, therefore, the model bias almost exists as origin, as we can see the trends of using OCG and no guidelines are relatively similar. These findings indicate that guidelines potentially improve context utilization and mitigate token bias.

\subsection{\model{} on Reasoning Tasks} \label{appx:longguide-in-reasoning}

\begin{wraptable}{c}{0.43\textwidth}
\footnotesize
\vspace{-3mm}
\scalebox{1}{
\begin{tabular}{lcc}
\toprule 
\textbf{Methods} & \textbf{GSM8k} & \textbf{SVAMP} \\
\midrule
Zero-shot (ZS) & 39.66 & 60.33 \\
+ APO & 41.83 & 62.33 \\
+ adv-ICL & 42.66 & 62.83 \\
\hdashline
+ \model{} & 40.83 & 63.33 \\
\midrule 
Few-shot (FS) & 32.33 & 61.66 \\
+ APO & 34.33 & 63.00 \\
+ adv-ICL & 35.00 & 62.66 \\
\hdashline
+ \model{} & 34.83 & 62.83 \\
\bottomrule
\end{tabular}
}
\caption{\small{Performance of \model{} with Mistral on reasoning tasks.}}
\vspace{-5mm}
\label{tab:longpo-on-reasoning}
\end{wraptable}

We conduct experiments comparing \model{} to various baselines on reasoning tasks. We select Mistral as our LLM, and GSM8K \citep{cobbe2021training} and SVAMP \citep{patel-etal-2021-nlp} as benchmarks for evaluation. For each benchmark, we randomly sampled 200 instances from the test set for assessment and 50 instances from the train set to train the prompt optimizers and \model{}. 

The results are averaged over three runs, and outlined in \Cref{tab:longpo-on-reasoning}. \model{} slightly outperforms the Zero-shot and Few-shot baselines but falls short compared to prompt optimizers. Nonetheless, the findings confirm that additional instructions for LLMs can potentially improve the init model, leading to further enhanced reasoning performance with prompt optimization.

\subsection{Supplementary Results for \Cref{sec:demos-anayn}} 

\begin{figure*}
\centering
\includegraphics[width=1\linewidth, trim={0cm 0cm 0cm 0cm},clip]{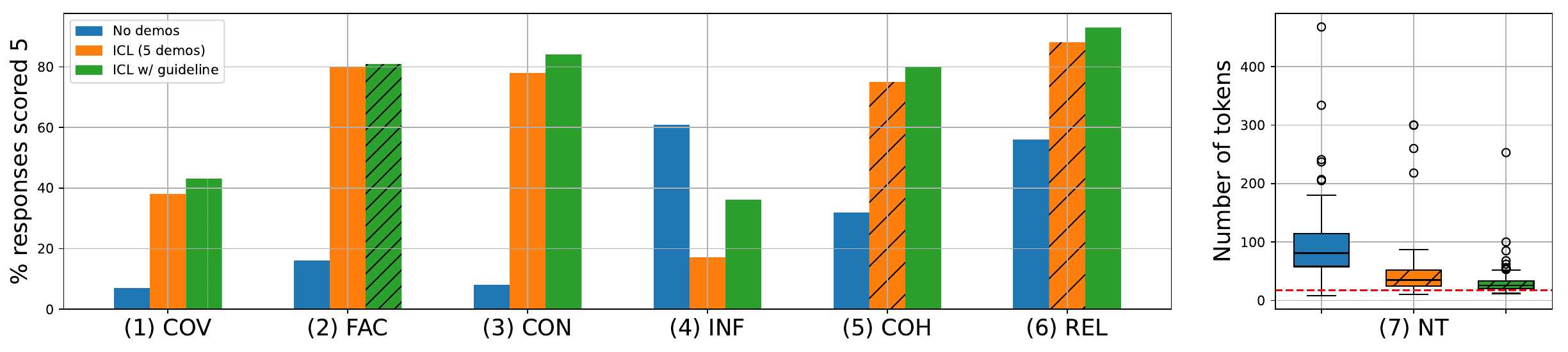}
\vspace{-7mm}
\caption{\small{Property maintenance experiments with ICL. See Appx.-\Cref{fig:full-example-theorem2.1} for a full example.}}
\label{fig:more-demos-does-not-help}
\vspace{-3mm}
\end{figure*}

\begin{figure}
\centering
\includegraphics[width=1\linewidth, trim={0cm 0cm 0cm 0cm},clip]{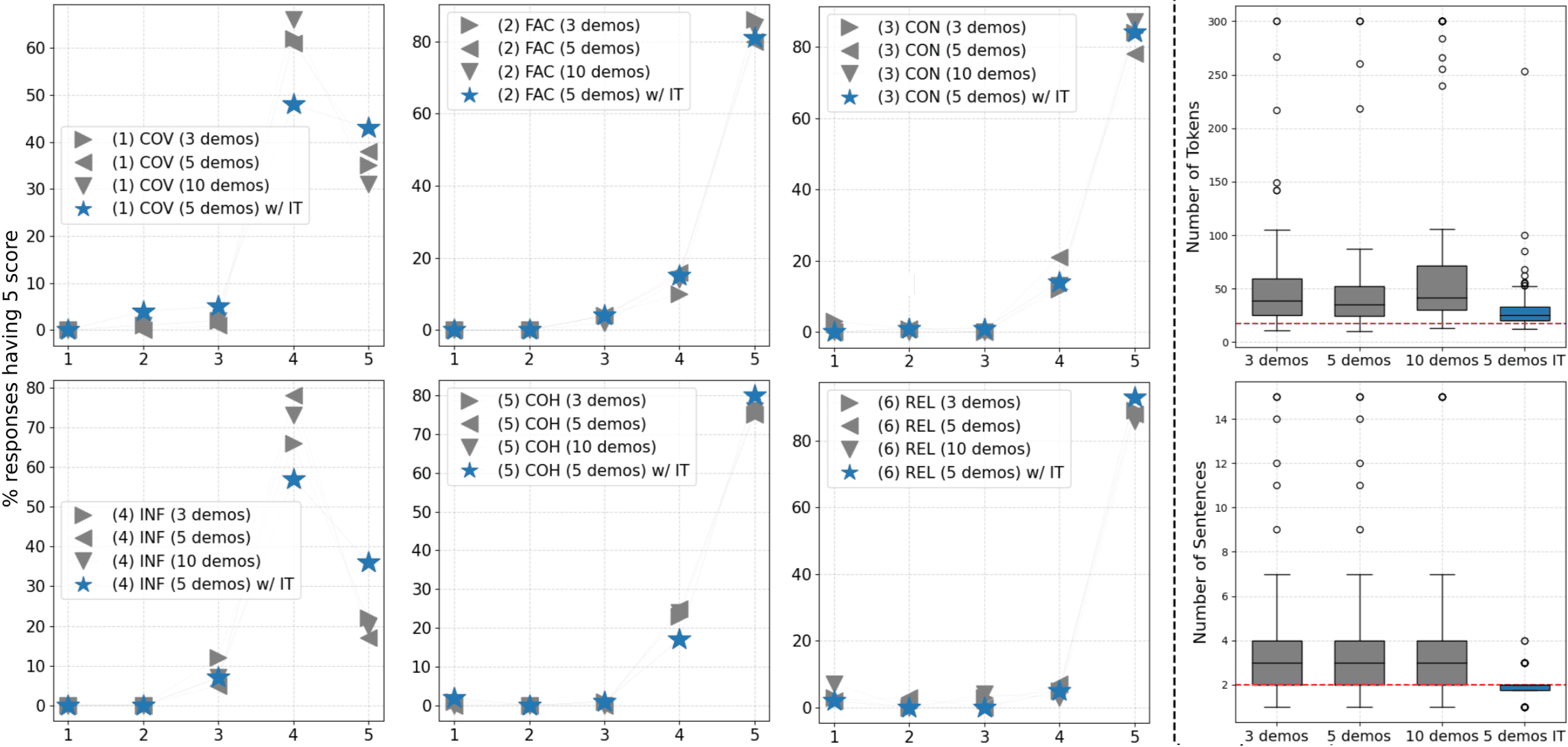}
\caption{\small{Property maintenance experiments with ICL full results. IT is the adding a simple guideline baseline.}}
\label{fig:more-demos-does-not-help-full-results}
\end{figure}

\begin{figure}
\small 
\centering
\vspace{-5mm}
\includegraphics[width=.3\linewidth, trim={0cm 0cm 0cm 0cm},clip]{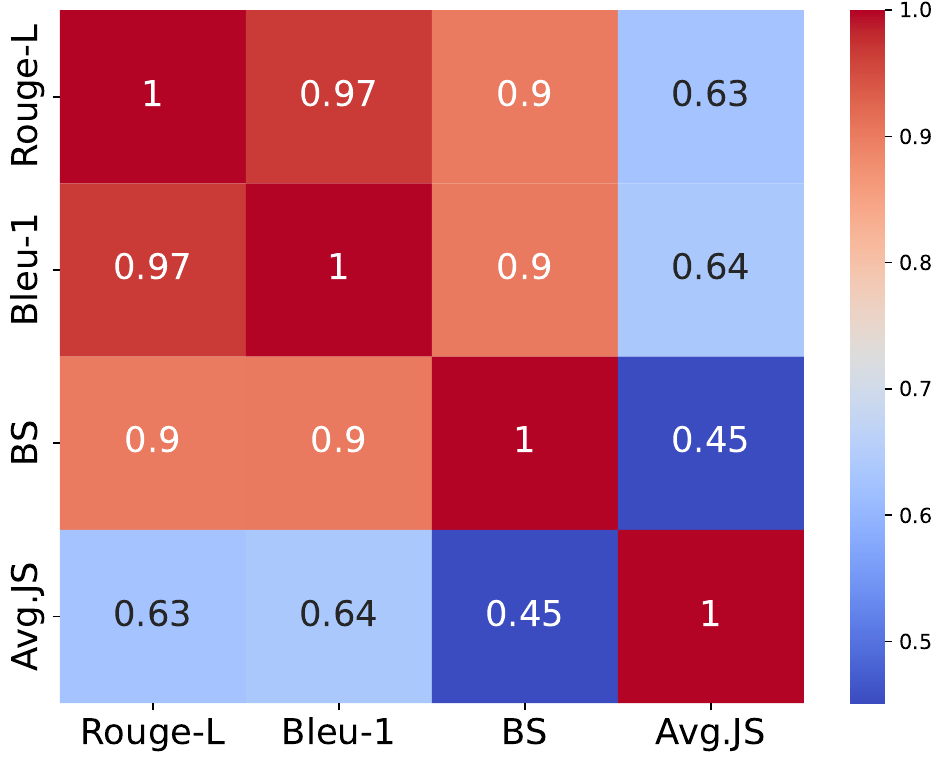}
\vspace{-3mm}
\caption{\small{Pairwise Pearson correlation coefficient of metrics.}}
\label{fig:pearson-correlation}
\end{figure}

\subsection{Understanding MG and OCG: How Do They Work (Together)?} \label{ssec:understanding-mg-ocg}

\begin{figure}
\centering
\includegraphics[width=1\linewidth, trim={0cm 0cm 0cm 0cm},clip]{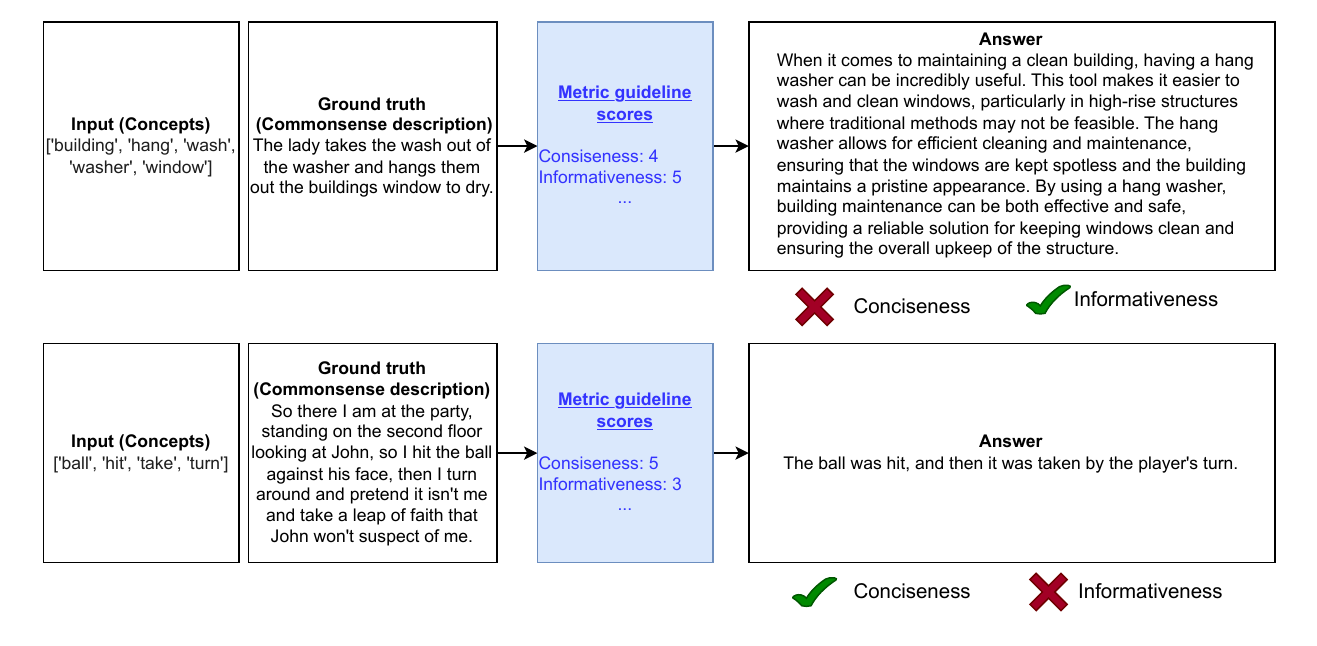}
\vspace{-6mm}
\caption{\small{A CommonGen-Challenge example \citep{lin-etal-2020-commongen}, where output with high Conciseness score could have low Informativeness score and vice versa}}\label{fig:conciseness-informativeness-example}
\vspace{-5mm}
\end{figure}

\paragraph{Metric Guideline (MG) (Step 1-3).} To understand how models select and evaluate metrics, we analyze the specific metrics chosen for each task, their selection frequencies, and their average scores (Appx.-\Cref{tab:selected-metrics-by-task,fig:metric-frequency,fig:metric-scores} respectively). Overall, each of the 27 metrics is selected and evaluated in at least one task. Among them, common linguistic metrics such as ``Clarity'' are frequently selected, while task-specific metrics like ``Creativity'' are less frequently chosen. By examining the scores of selected metrics, we find that common linguistic metrics generally achieve high scores, as anticipated. However, task-specific metrics like ``Creativity'' exhibit varying scores across tasks, indicating their differing importance and relevance. Additionally, we also find that within MG can conflict with each other, such as ``Conciseness'' and ``Informativeness'' (see Appx.-\Cref{fig:conciseness-informativeness-example} for an example). This underscores the importance of \model{}'s Step 2 in weighting the metrics.

\paragraph{Output Constraint Guideline (OCG) (Step 4).}
We find that both the token and sentence constraints are crucial for LLMs (Appx.-\ref{appdx:extra-ablations}), with the sentence being more beneficial. We hypothesize that LLMs have better control over the number of sentences than tokens, as counting sentences is intuitively simpler than tokens. This can be observed in our experiment in \Cref{sec:demos-anayn}.

\begin{figure}
\centering
\includegraphics[width=.95\linewidth, trim={0cm 0cm 0cm 0cm},clip]{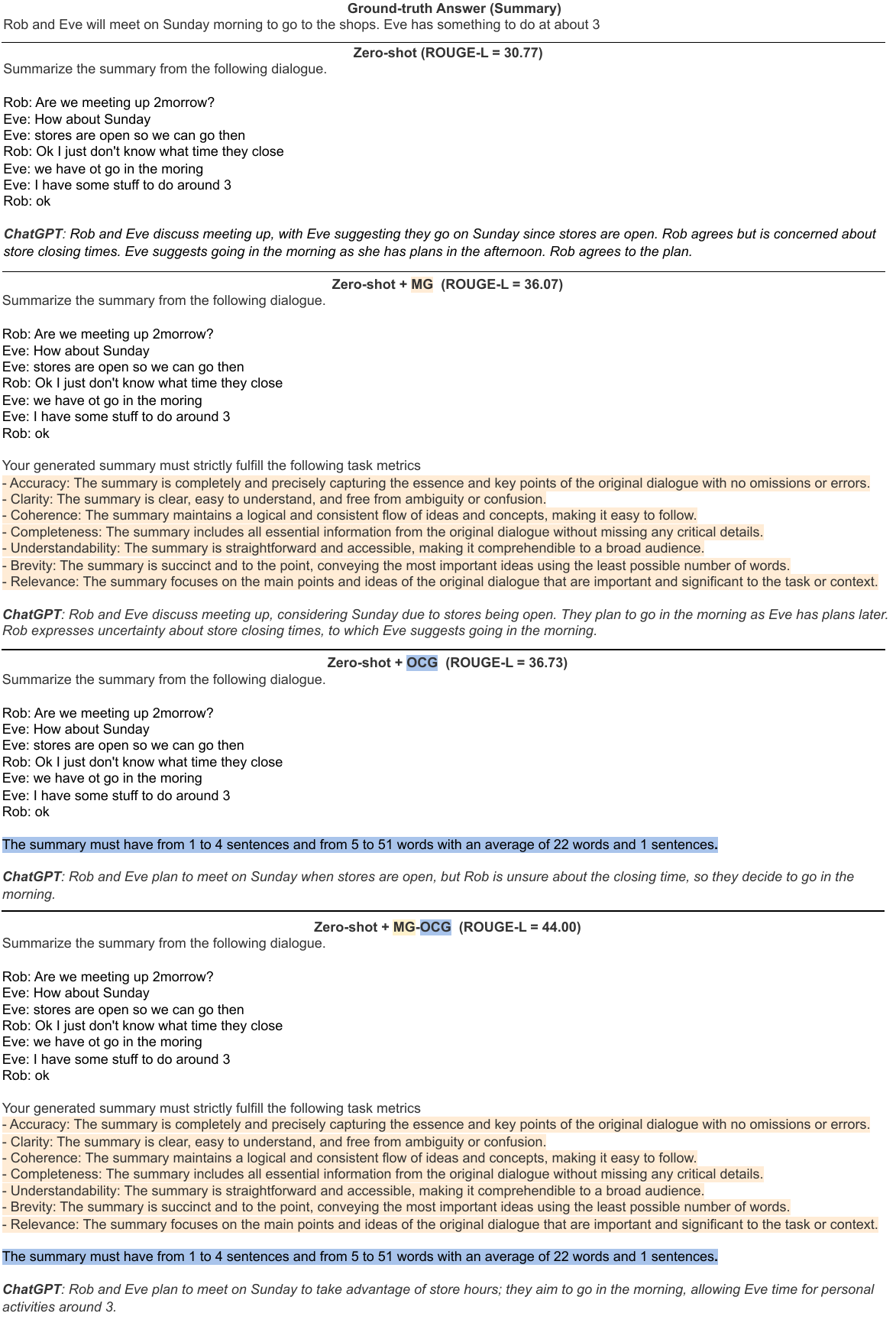}
\caption{\small{A sample from SAMSum dataset where MG and OCG supplement each other and are not interchangeable to increase the performance in final answer.}}\label{fig:MG-OCG-supplement}
\end{figure}

\begin{figure}
\centering
\includegraphics[width=\linewidth, trim={0cm 0cm 0cm 0cm},clip]{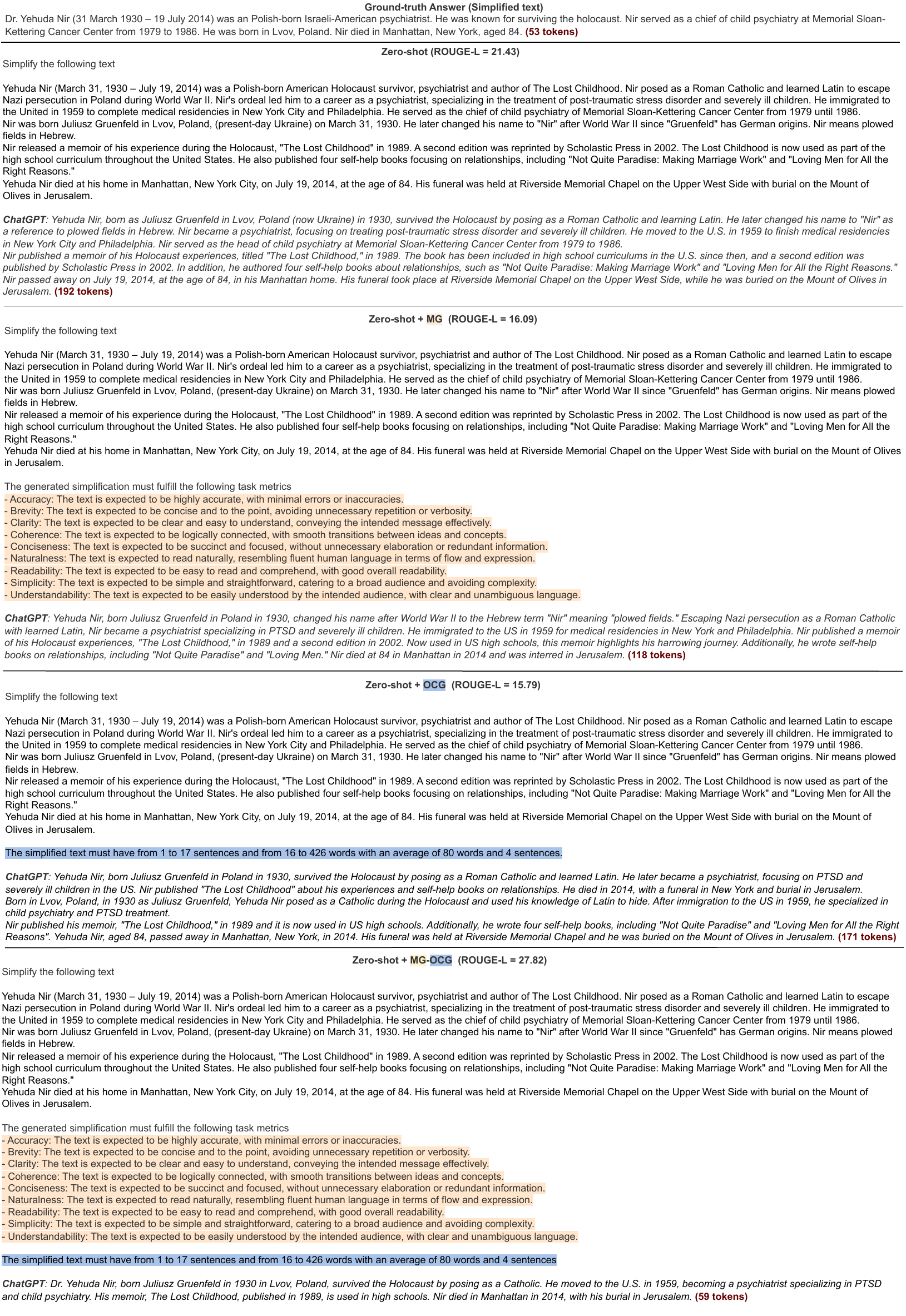}
\vspace{-3mm}
\caption{\small{An example of SWiPE \citep{laban-etal-2023-swipe} where the record contains fewer tokens than the expected average. This reduces the effectiveness of OCG and MG individually, but their combination could enhance performance.}}
\label{fig:out-of-distribution-example}
\vspace{-5mm}
\end{figure}

\paragraph{MG and OCG are complementary and non-interchangeable.} 
MG and OCG complement each other rather than conflict, as partially discussed in \Cref{ssec:main-findings}. This is because MG language metrics primarily concern the characteristics of responses rather than their structural aspects such as sentence and token count, which is the main focus of the OCG. In addition, the MG and OCG are not interchangeable. One might question whether adopting conciseness and brevity metrics could sufficiently alter the OCG, or if the OCG could effectively encompass the MG guideline. Our answer is no. While MG can steer LLMs towards brevity in responses, it lacks precise quantification for conciseness. Modern LLMs, often trained to generate verbose responses, may struggle to meet human conciseness without explicit statistics. Meanwhile, the OCG supplies them in the form of bins and means, yet these statistics alone do not directly address linguistic qualities. We provide examples as evidence supporting our arguments in Appx.-\Cref{fig:MG-OCG-supplement,fig:out-of-distribution-example}.

\subsection{Collected Metrics in \model{}'s Step 1 (\Cref{sec:method})}

\Cref{tab:evaluation-metrics} presents our $27$ metrics collected for \model{}'s Step 1.

\begin{table*}
\centering
\scalebox{.7}{
\begin{tabular}{l|l|c}
\toprule
\textbf{Source}  & \textbf{Metrics} & \textbf{\#} \\
\midrule
The ABC's of Communication \citep{wagner1963abc} & Accuracy, Brevity, Clarity & 3 \\
\midrule
BARTScore \citep{yuan2021bartscore} & Relevance, Coherence & 2 \\
\midrule
GPTScore \citep{fu2023gptscore} & Semantic Coverage, Factuality, Fluency, Informativeness, & 10 \\
& Consistency, Engagement, Specificity, Correctness,  & \\
& Understandability, Diversity & \\
\midrule
We propose & Completeness, Conciseness, Neutrality, Naturalness, Readability, Creativity, & 12 \\
& Rationalness, Truthfulness, Respect of Chronology, &  \\
&  Non-repetitiveness, Indicativeness, Resolution &  \\
\midrule
\textbf{Total} & 27 & 27 \\
\bottomrule
\end{tabular}
}
\caption{\small{Metrics collected for \model{}'s metric guideline (MG).}}
\label{tab:evaluation-metrics}
\end{table*}

\subsection{JS Divergence over all \model{} Metrics with SAMSum (\Cref{ssec:main-findings})} \label{appdx:js-components}

\begin{figure*}[hp!]
\includegraphics[width=\linewidth, trim={0cm 0cm 0cm 0cm},clip]{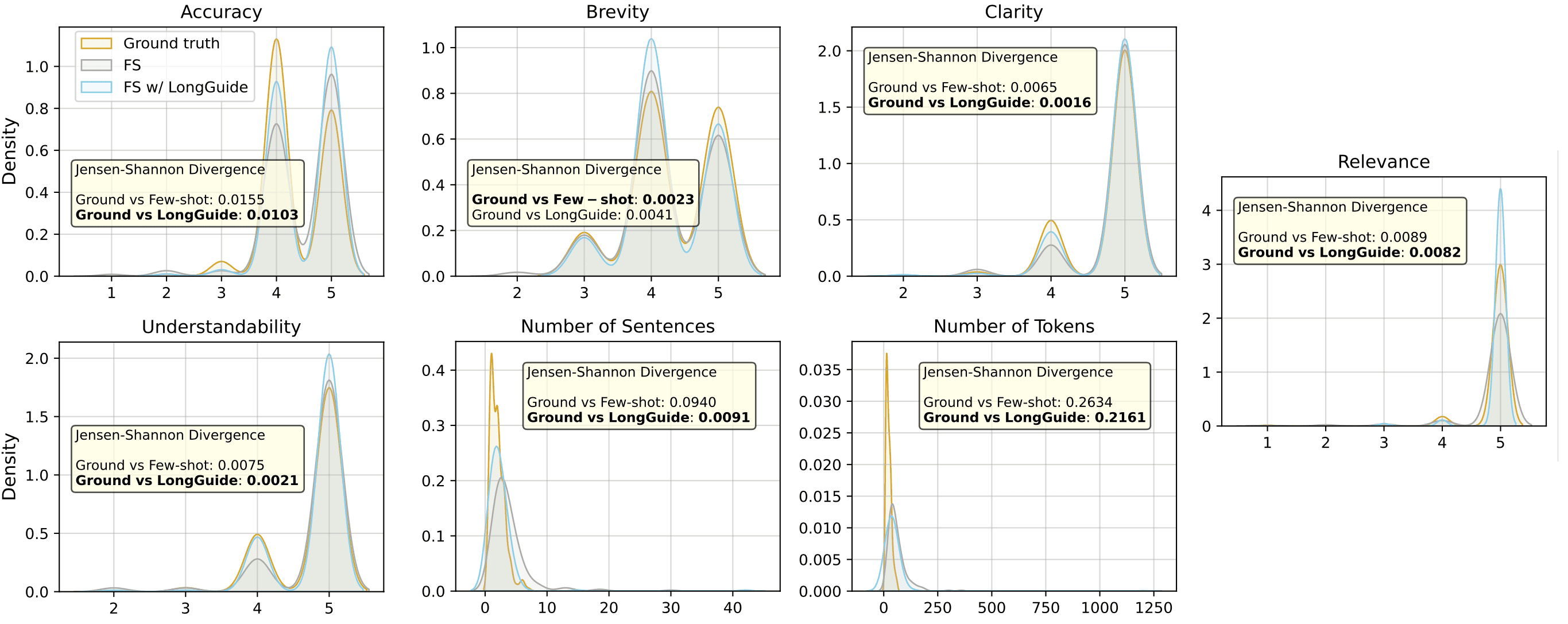}
\caption{\small{Density plots of MG and OCG metrics selected by Mistral under the few-shot (FS) setting, measured on ground-truth, FS, and FS w/ \model{} answers. For Jensen–Shannon divergence, \textbf{lower is better}.}}
\label{fig:how-mg-ocg-work}
\end{figure*}

\Cref{fig:how-mg-ocg-work} presents density plots of MG and OCG metrics selected by Mistral under the few-shot (FS) setting, measured on ground-truth, FS, and FS w/ \model{} answers. For Jensen–Shannon divergence, the lower is better.

\subsection{Step 5 CD-MG Selection Results of \model{} (\Cref{ssec:main-findings})}  

\begin{table}[hp!]
\centering
\footnotesize
\scalebox{.67}{
\begin{tabular}{cl|ccc|c|c|c|c}
\toprule
 & &  & \textbf{Summarization} & & \textbf{Simplification} & \textbf{Translation} & \textbf{Dialogue Generation}  &\textbf{Table2Text} \\
\midrule
& \textbf{Method} & \textbf{SAMSum} & \textbf{CNN (3.0.0)} & \textbf{XL-Sum} & \textbf{SWiPE} & \textbf{IWSLT17 en-ja} & \textbf{Synthetic Persona} & \textbf{CommGen-Chall.} \\
\midrule
& \textbf{\#shots (random)} &  3 & 3 & 5 & 3 & 5 &  5 & 5 \\
\midrule
\multirow{10}*{\begin{tabular}[c]{@{}l@{}}{\rotatebox{90}{Mistral-7B-it}}
\end{tabular}} 
& Zero-shot (ZS) & 21.25  & 18.96 & 8.88 & 36.21 & 14.05 & 12.93 & 9.12  \\
& \emph{+ OCG} & 27.43 & \textbf{21.92} & 
\textbf{14.22} & 31.19 & \textbf{16.93} & 12.99 & 20.67 \\
& \emph{+ MG} & 27.68 & 18.02 & 10.26 & \textbf{36.74} & 11.06 & 13.74 & 19.98 \\
& \emph{+ MG-OCG} & \textbf{28.34} & 21.63 & 13.90 & 35.12 & 15.49 & \textbf{14.14} & \textbf{20.87}  \\
& \textcolor{gray}{\emph{MG-OCG Sel.}} & \textcolor{gray}{\emph{MG-OCG}} & \textcolor{gray}{\emph{OCG}} & \textcolor{gray}{\emph{OCG}} & \textcolor{gray}{\emph{MG}} & \textcolor{gray}{\emph{OCG}} & \textcolor{gray}{\emph{MG-OCG}} & \textcolor{gray}{\emph{MG-OCG}} \\
\cmidrule{2-9}
& Few-shot (FS)  & 25.55 & 17.30 & 9.85 & 39.29 & 13.52 & 6.19 & 4.01 \\
& \emph{+ OCG} & 27.31 & 16.45 & 12.47 & 29.85 & \textbf{17.58} & 6.45 & \textbf{20.50} \\
& \emph{+ MG} & 27.88 & 18.47 & 12.01 & \textbf{41.07} & 14.09 & 6.47 & 11.16 \\
& \emph{+ MG-OCG} & \textbf{30.01} & \textbf{19.87} & \textbf{14.89} & 39.40 & 17.02 & \textbf{8.06} & 5.18 \\
& \textcolor{gray}{\emph{MG-OCG Sel.}} & \textcolor{gray}{\emph{MG-OCG}} & \textcolor{gray}{\emph{MG-OCG}} & \textcolor{gray}{\emph{MG-OCG}} & \textcolor{gray}{\emph{MG}} & \textcolor{gray}{\emph{OCG}} & \textcolor{gray}{\emph{MG-OCG}} & \textcolor{gray}{\emph{OCG}} \\
\midrule
\multirow{10}*{\begin{tabular}[c]{@{}l@{}} {\rotatebox{90}{ChatGPT}} \\
\end{tabular}} 
& Zero-shot (ZS) & 24.21 & 19.54 & 10.78 & \textbf{45.11} & 36.22 & 19.68 & 24.23 \\
& \emph{+ OCG} & 28.81 & 21.88 & \textbf{20.66} & 37.58 & 38.45 & \textbf{23.09} & \textbf{35.04} \\
& \emph{+ MG} & 25.12 & 20.02 & 10.42 & 45.09 & 37.72 & 19.81 & 18.50 \\
& \emph{+ MG-OCG} & \textbf{29.79} & \textbf{21.99} & 19.91 & 42.72 & \textbf{41.50} & 20.82 & 30.09\\
& \textcolor{gray}{\emph{MG-OCG Sel.}} & \textcolor{gray}{\emph{MG-OCG}} & \textcolor{gray}{\emph{MG-OCG}} & \textcolor{gray}{\emph{OCG}} & \textcolor{gray}{\emph{ZS}} & \textcolor{gray}{\emph{MG-OCG}} & \textcolor{gray}{\emph{MG-OCG}} & \textcolor{gray}{\emph{OCG}} \\
\cmidrule{2-9}
& Few-shot (FS)  & 27.44 & 13.77 & 12.11 & 33.30 & 28.76 & 17.12 & 24.12 \\
& \emph{+ OCG} & 29.98 & \textbf{17.55} & \textbf{19.26} & 16.22 & 35.73 & \textbf{21.50} & 36.51 \\
& \emph{+ MG} & 28.89 & 14.03 & 12.75 & 19.14 & 36.09 & 19.12 & 21.99 \\
& \emph{+ MG-OCG} & \textbf{30.65} & 13.12 & 18.64 & \textbf{37.24} & \textbf{36.22} & 18.99 & \textbf{38.33} \\
& \textcolor{gray}{\emph{MG-OCG Sel.}} & \textcolor{gray}{\emph{MG-OCG}} & \textcolor{gray}{\emph{OCG}} & \textcolor{gray}{\emph{OCG}} & \textcolor{gray}{\emph{MG-OCG}} & \textcolor{gray}{\emph{MG-OCG}} & \textcolor{gray}{\emph{OCG}} & \textcolor{gray}{\emph{MG-OCG}} \\
\bottomrule
\end{tabular}}
\caption{
\small{MG-OCG selection results on $D^{train}$ set for the main experiments in \Cref{table:main-experimental-results}, evaluated by ROUGE-L.} 
}
\label{table:CD-MG-selection-results}
\end{table}

The numerical MG-OCG selection results on $D^{train}$ are presented in \Cref{table:CD-MG-selection-results}, as also noted in \Cref{table:main-ablation-results}. Overall, the performance of \model{} on $D^{train}$ closely mirrors its performance on the testing tasks in \Cref{table:main-ablation-results}. The only discrepancy is for the IWSLT17 en-ja task with ChatGPT using few-shot prompting: the optimal guideline combination on $D^{train}$ is MG-OCG (see \Cref{table:CD-MG-selection-results}), whereas the best on the testing set is MG (see \Cref{table:main-ablation-results}).

\subsection{\model{} can Generalize from Demonstrations (\Cref{ssec:generalize-from-demos})}

\begin{table}
\centering
\footnotesize
\scalebox{1}{
\begin{tabular}{lccc}
\toprule 
\textbf{Methods} & \textbf{CNN (3.0.0)} & \textbf{IWSLT17 en-ja} & \textbf{CommGen-Chall.}\\
\midrule
Zero-shot (ZS) & 19.23$_{\pm 0.34}$ & 13.12$_{\pm 1.39}$ & 10.12$_{\pm 0.02}$ \\
+ OCG trained on $D^{train}$ & \textbf{22.46}$_{\pm 0.64}$ & 16.53$_{\pm 0.59}$ & 24.16$_{\pm0.11}$ \\
+ MG trained on $D^{train}$ & 18.35$_{\pm 0.60}$ & 8.71$_{\pm 0.53}$ & 21.54$_{\pm 7.50}$ \\
+ MG-OCG trained on $D^{train}$ & 22.05$_{\pm 0.84}$ & 15.76$_{\pm 1.85}$ & \textbf{25.20}$_{\pm 1.89}$ \\
+ \model{} trained on $D^{train}$ & \textbf{22.46}$_{\pm 0.64}$ & 16.53$_{\pm 0.59}$ & \textbf{25.20}$_{\pm 1.89}$ \\
\hdashline
+ OCG trained on Demos & 20.46$_{\pm 0.10}$ & \textbf{17.27}$_{\pm 1.83}$  & 23.97$_{\pm 0.47}$ \\
+ MG trained on Demos & 18.33$_{\pm 0.25}$ & 8.63$_{\pm 1.08}$  & 18.98$_{\pm 0.52}$ \\
+ MG-OCG trained on Demos & 19.16$_{\pm 0.37}$ & 14.00$_{\pm 3.42}$  & 24.46$_{\pm 2.43}$ \\
+ \model{} trained on Demos & 20.46$_{\pm 0.10}$ & 14.00$_{\pm 2.42}$  & 24.46$_{\pm 2.43}$ \\
\midrule
Few-shot (FS) & 17.56$_{\pm 0.63}$ & 12.69$_{\pm 1.82}$ & 3.98$_{\pm0.17}$ \\
+ OCG trained on $D^{train}$ & 19.17$_{\pm 1.27}$ & \textbf{19.86}$_{\pm 2.93}$ & 25.05$_{\pm 0.76}$ \\
+ MG trained on $D^{train}$ & 17.18$_{\pm 2.01}$ & 12.82$_{\pm 0.15}$ & 21.79$_{\pm 5.20}$ \\
+ MG-OCG trained on $D^{train}$ & \textbf{21.18}$_{\pm 1.07}$ & 18.70$_{\pm 0.73}$ & 25.43$_{\pm 5.28}$ \\
+ \model{} trained on $D^{train}$ & \textbf{21.18}$_{\pm 1.07}$ & \textbf{19.86}$_{\pm 2.93}$ & 25.05$_{\pm 0.76}$ \\
\hdashline
+ OCG trained on Demos & 16.88$_{\pm 1.44}$ & 19.40$_{\pm 1.39}$  & \textbf{28.28}$_{\pm 0.69}$ \\
+ MG trained on Demos & 15.59$_{\pm 0.59}$ & 12.07$_{\pm 2.68}$  & 23.99$_{\pm 4.66}$ \\
+ MG-OCG trained on Demos & 19.89$_{\pm 0.39}$ & 17.78$_{\pm 3.23}$  & 27.41$_{\pm 0.87}$ \\
+ \model{} trained on Demos & 19.89$_{\pm 0.39}$ & 17.78$_{\pm 18.43}$  & 23.99$_{\pm 4.66}$ \\
\bottomrule
\end{tabular}
}
\caption{\small{\model{} learns the guidelines from only demonstrations with Mistral, evaluated by ROUGE-L.}}
\label{tab:longpo-learns-from-demos}
\end{table}

\Cref{tab:longpo-learns-from-demos} presents the numerical results of \Cref{fig:longguide-learns-from-demos} in \Cref{ssec:generalize-from-demos}. Even with only 3-5 exemplars as demonstrations, \model{} effectively derives MG and OCG guidelines, benefiting the model. In this case, $D^{train}$ is the set of demonstrations, and the rest of \model{}'s steps remain unchanged.

\subsection{Human Evaluation Fine-grained Results (\Cref{ssec:human-eval})} \label{ssec:human-evaluation-rating}

\begin{figure*}[hp!]
\includegraphics[width=\linewidth, trim={0cm 0cm 0cm 0cm},clip]{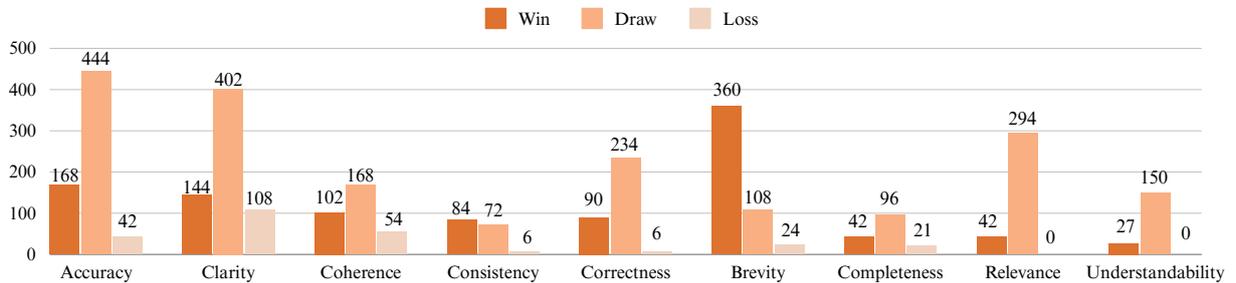}
\caption{\small{Fine-grained human evaluation results on evaluated MG metrics.}}
\label{fig:finegrained-human-eval}
\end{figure*}

\Cref{fig:finegrained-human-eval} presents our fine-grained human evaluation results. Overall, \model{} shows the best in terms of ``Accuracy'' and ``Clarity'', with a significant number of winning ratings. This suggests that the generated text is factually correct and easy to understand. Meanwhile, \model{} shows more mixed results in terms of ``Clarity'' and ``Coherence''. While there is still a high winning rating, the proportion of draw and loss ratings is also relatively high, possibly because improving ``Brevity'' can somehow reduce the ``Clarity''. 

\subsection{Attention Analysis for Guideline Tokens (\Cref{ssec:attention-analysis})}

\begin{table}
\centering
\resizebox{0.9\textwidth}{!}{ 
\begin{tabular}{l|cc|cc|cc} 
\hline
 & MG & Context (MG) & OCG & Context (OCG) & MG-OCG & Context (MG-OCG) \\ \hline
Per token & 0.0019 & 0.0064 & 0.0133 & 0.0077 & 0.0017 & 0.0064 \\ \hline
All & 37.81\% & 62.19\% & 22.56\% & 77.44\% & 37.87\% & 62.13\% \\ \hline
\end{tabular}}
\caption{Attention score over guideline and context tokens of Mistral.}
\label{tab:attention-score-analysis}
\end{table}

\Cref{tab:attention-score-analysis} shows our simple attention analysis.
\begin{table}[hp!]
\centering
\scalebox{0.75}{ 
\renewcommand{\arraystretch}{0.9} 
\begin{tabular}{p{2.5cm}|p{2cm}|p{15cm}} 
\toprule
\textbf{Task} & \textbf{Model} & \textbf{Selected Metrics}  \\
\midrule
SAMSum & Mistral & ['Accuracy', 'Brevity', 'Clarity', 'Relevance', 'Understandability'] \\
 & ChatGPT & ['Accuracy', 'Brevity', 'Clarity', 'Relevance', 'Understandability'] \\
\midrule
CNN & Mistral & ['Accuracy', 'Brevity', 'Clarity', 'Coherence', 'Completeness', 'Engagement', 'Readability', 'Relevance', 'Truthfulness', 'Understandability'] \\
& ChatGPT & ['Accuracy', 'Brevity', 'Clarity', 'Coherence', 'Completeness', 'Conciseness', 'Engagement', 'Neutrality', 'Readability', 'Relevance', 'Specificity'] \\
\midrule
XLSum & Mistral & ['Accuracy', 'Brevity', 'Clarity', 'Coherence', 'Completeness', 'Consistency', 'Correctness', 'Diversity', 'Engagement', 'Factuality', 'Fluency', 'Indicative', 'Informativeness', 'Neutrality', 'Non-repetitiveness', 'Relevance', 'Resolution', 'Respect of Chronology', 'Semantic Coverage', 'Specificity', 'Understandability'] \\
& ChatGPT & ['Accuracy', 'Brevity', 'Clarity', 'Coherence', 'Completeness', 'Consistency', 'Correctness', 'Diversity', 'Engagement', 'Factuality', 'Fluency', 'Indicative', 'Informativeness', 'Neutrality', 'Non-repetitiveness', 'Rationalness',  'Relevance', 'Resolution', 'Respect of Chronology', 'Semantic Coverage', 'Specificity', 'Understandability'] \\
\midrule
SWiPE & Mistral & ['Accuracy', 'Brevity', 'Clarity', 'Relevance', 'Understandability'] \\
 & ChatGPT & ['Accuracy', 'Brevity', 'Clarity', 'Coherence', 'Conciseness', 'Consistency', 'Correctness', 'Readability', 'Understandability'] \\
\midrule
IWSLT17 en-ja & Mistral & ['Accuracy', 'Clarity', 'Coherence', 'Consistency', 'Correctness', 'Factuality', 'Fluency', 'Relevance', 'Understandability'] \\
& ChatGPT & ['Accuracy', 'Clarity', 'Coherence', 'Consistency', 'Correctness', 'Factuality', 'Fluency', 'Relevance', 'Understandability'] \\
\midrule
Synthetic Persona & Mistral & ['Accuracy', 'Brevity', 'Clarity', 'Coherence', 'Completeness', 'Consistency', 'Correctness', 'Diversity', 'Engagement', 'Factuality', 'Fluency', 'Indicative', 'Informativeness', 'Neutrality', 'Non-repetitiveness', 'Relevance', 'Resolution', 'Respect of Chronology', 'Semantic Coverage', 'Specificity', 'Understandability'] \\
& ChatGPT & ['Accuracy', 'Clarity', 'Coherence', 'Consistency', 'Correctness', 'Diversity', 'Engagement', 'Fluency', 'Indicative', 'Informativeness', 'Neutrality', 'Non-repetitiveness', 'Relevance', 'Resolution', 'Respect of Chronology', 'Specificity', 'Understandability'] \\
\midrule
CommGen-Chall. & Mistral & ['Coherence', 'Conciseness', 'Fluency', 'Relevance', 'Understandability'] \\
& ChatGPT & ['Clarity', 'Coherence', 'Completeness', 'Conciseness', 'Consistency', 'Creativity', 'Engagement', 'Fluency', 'Naturalness', 'Relevance'] \\
\bottomrule
\end{tabular}
}
\caption{\small{Selected metrics by tasks by Mistral and ChatGPT.}}
\label{tab:selected-metrics-by-task}
\end{table}

\subsection{Which Metrics Were Selected the Most for MG? (\Cref{ssec:understanding-mg-ocg})} 

\begin{figure}
\centering
\begin{subfigure}[t]{\linewidth} 
\centering
\includegraphics[width=.9\linewidth, trim={0cm 0cm 0cm 0cm}, clip]{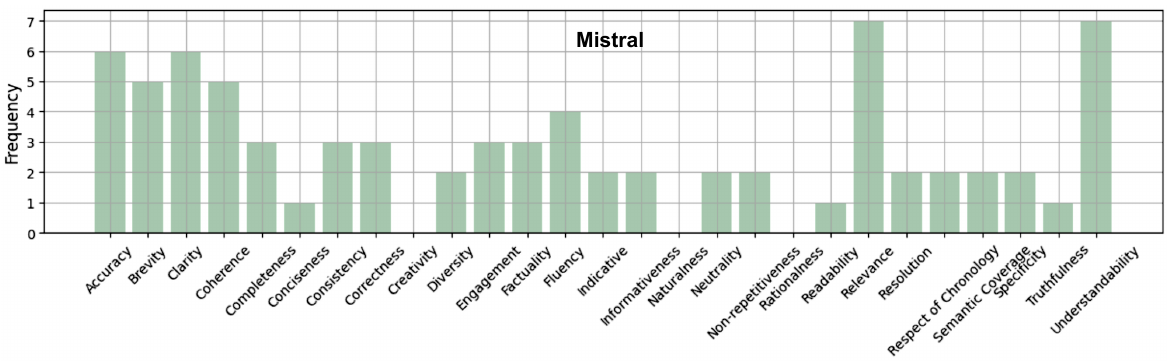}
\end{subfigure}
\vspace{5mm} 
\begin{subfigure}[b]{\linewidth} 
\centering
\includegraphics[width=.9\linewidth, trim={0cm 0cm 0cm 0cm}, clip]{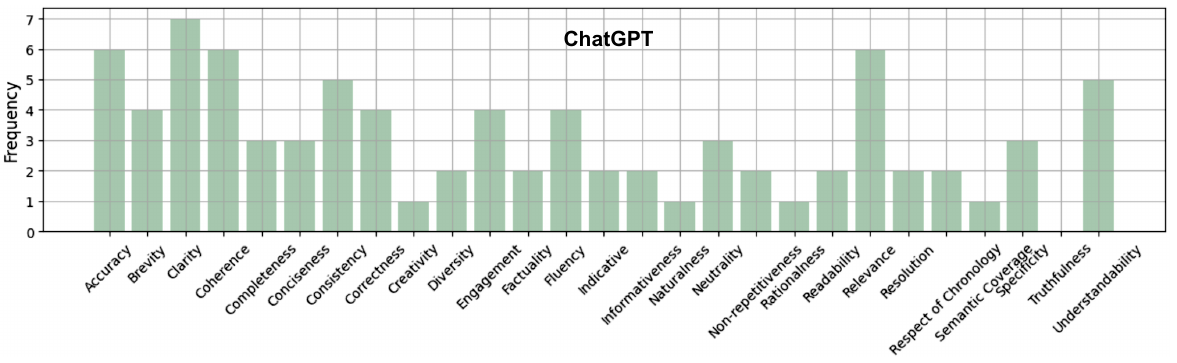} 
\vspace{-10mm}
\end{subfigure}
\caption{\small Frequency of metrics selected as the metric guideline.} 
\label{fig:metric-frequency}
\end{figure}

\begin{figure}
\centering
\begin{subfigure}[t]{\linewidth} 
\centering
\includegraphics[width=.9\linewidth, trim={0cm 0cm 0cm 0cm}, clip]{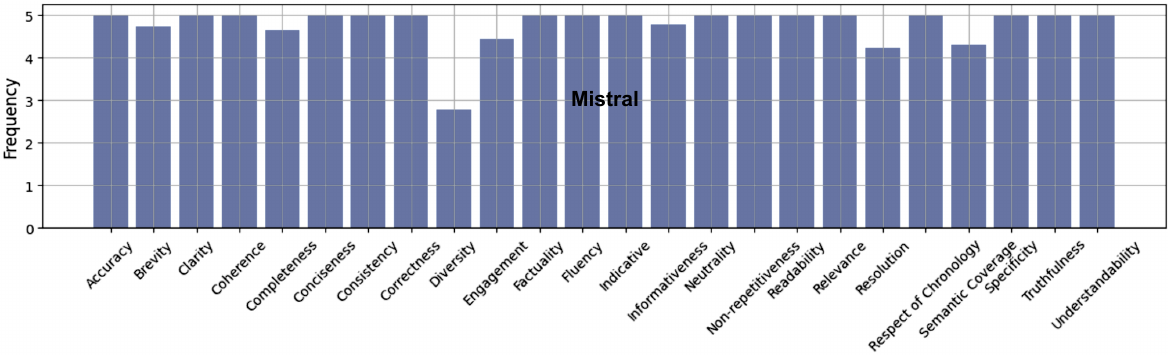}
\end{subfigure}
\vspace{5mm} 
\begin{subfigure}[b]{\linewidth} 
\centering
\includegraphics[width=.9\linewidth, trim={0cm 0cm 0cm 0cm}, clip]{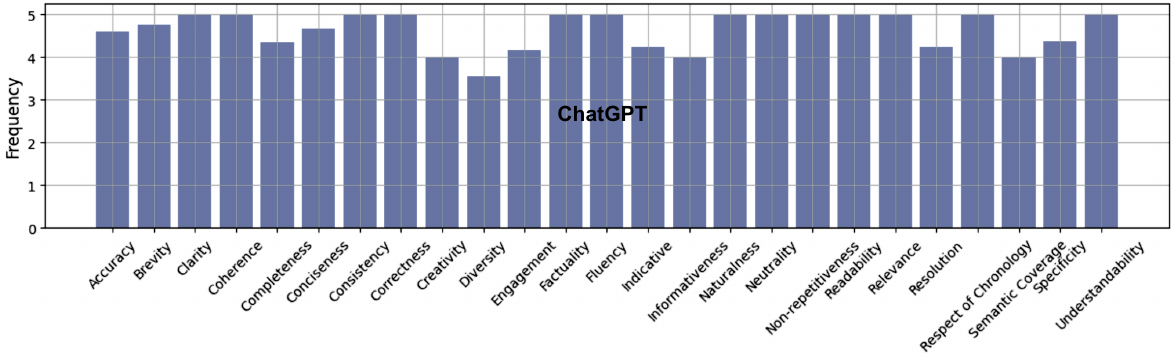} 
\vspace{-10mm}
\end{subfigure}
\caption{\small Average scores of metrics as the metric guideline.} 
\vspace{-3mm}
\label{fig:metric-scores}
\end{figure}

To better understand how models select and evaluate metrics, we analyze the specific metrics chosen for each task (\Cref{tab:selected-metrics-by-task}), their selection frequencies (\Cref{fig:metric-frequency}), and their average scores (\Cref{fig:metric-scores}).

\subsection{Extra Ablation Studies: without OCG's Token or Sentence Constraint (\Cref{ssec:understanding-mg-ocg})}\label{appdx:extra-ablations}

\begin{table}[hp!]
\small 
\centering
\begin{tabular}{lccc}
\toprule 
\textbf{Methods} & \textbf{CNN (3.0.0)} & \textbf{IWSLT17 en-ja} & \textbf{CommGen-Chall.}\\
\midrule
Zero-shot (ZS) & 19.23$_{\pm 0.34}$ & 13.12$_{\pm 1.39}$ & 10.12$_{\pm 0.02}$ \\
+ \model{}  & \textbf{22.46}$_{\pm 0.64}$ & \textbf{16.53}$_{\pm 0.59}$ & \textbf{25.20}$_{\pm 1.89}$ \\
+ \model{} w/o Token Constraint & 21.54$_{\pm 0.52}$\textcolor[RGB]{255,0,0}{$\downarrow$} & 14.09$_{\pm 1.07}$\textcolor[RGB]{255,0,0}{$\downarrow$} & 21.49$_{\pm 2.15}$\textcolor[RGB]{255,0,0}{$\downarrow$} \\
+ \model{} w/o Sentence Constraint & 20.92$_{\pm 0.23}$\textcolor[RGB]{255,0,0}{$\downarrow$} & 10.02$_{\pm 4.17}$\textcolor[RGB]{255,0,0}{$\downarrow$} & 13.32$_{\pm 0.73}$\textcolor[RGB]{255,0,0}{$\downarrow$} \\
\midrule
\midrule
Few-shot (FS) & 17.56$_{\pm 0.63}$ & 12.69$_{\pm 1.82}$ & 3.98$_{\pm0.17}$ \\
+ \model{} & \textbf{21.18}$_{\pm 1.07}$ & \textbf{19.86}$_{\pm 2.93}$ & \textbf{25.05}$_{\pm 0.76}$ \\
+ \model{} w/o Token Constraint & 20.30$_{\pm 1.46}$\textcolor[RGB]{255,0,0}{$\downarrow$} & 19.75$_{\pm 1.47}$\textcolor[RGB]{255,0,0}{$\downarrow$} & 20.30$_{\pm 1.46}$\textcolor[RGB]{255,0,0}{$\downarrow$} \\
+ \model{} w/o Sentence Constraint & 15.89$_{\pm 2.26}$\textcolor[RGB]{255,0,0}{$\downarrow$} & 12.57$_{\pm 2.99}$\textcolor[RGB]{255,0,0}{$\downarrow$}  & 12.20$_{\pm 3.91}$\textcolor[RGB]{255,0,0}{$\downarrow$} \\
\bottomrule
\end{tabular}
\caption{\small{Mistral results when omitting OCG's Token or Sentence Information, showing the importance of OCG's Token and Sentence information, evaluated by ROUGE-L.}
}
\label{tab:extra-ablation-ocg}
\end{table}

Since OCG's token information and sentence information are the two types of information emphasized in OCG, we further investigate the importance of each type of information. The empirical experiments are conducted with Mistral on CNN, IWSLT-2017 en-ja, and CommonGen-Challenge. We present the results in \Cref{tab:extra-ablation-ocg}. We observe that skipping OCG's token information or sentence information would hurt the performance. Specifically, the results drop more significantly when sentence information is omitted, and even fall below the Zero-shot score in CNN Few-shot with LongGuide and IWSLT17 en-ja Few-shot with LongGuide. The performance drops significantly in the CommonGen-Challenge Few-shot case, with a fall of 55.20\%. Due to the volatility of the token count in a sentence, it is hard to estimate the other information with only one type of information given. Therefore, both types of information should be provided to better capture the text distribution.

\section{Implementation Details}
\label{Appx:baselines-details}

\paragraph{Task benchmark preprocessing.} We chose the newest versions of the above datasets. For each dataset except Synthetic-Persona-Chat, we sample $200$ samples from the test set for our evaluation, following \citet{bai2023longbench}, and $50$ random samples from the train set for $D^{train}$. For Synthetic-Persona-Chat, we randomly sample $25$ dialogues from its test set for our evaluation ($678$ utterances in total) and $3$ dialogues from its train set where $50$ random utterances are selected for $D^{train}$.

\paragraph{Prompting baselines' hyperparameters.} We present the implementation and hyperparameters' details for our proposed \model{} as well as prompting baselines below.

\begin{itemize}
  \item \textbf{\model{}.} We set the batch size is $5$ and number of iterations is also $5$ for \model{}'s step 1. For steps 2, 3, and 4, no hyperparameter involves. For the evaluations by Self-consistency \citep{wang2022self}, we sample $3$ results. 
  \item \textbf{APO \citep{pryzant-etal-2023-automatic}.} We set the number of optimization iterations is $5$. We use $1$ sample with the lowest ROUGE-L score as the error sample for generating gradients, following \cite{do2024prompt}. At each iteration, $5$ textual gradients are generated, and $5$ new prompts are sampled from textual gradients. Finally, $1$ paraphrase of the input prompt is sampled at each optimization iteration. 
  \item \textbf{adv-ICL \citep{do2024prompt}.} We use $3$ iterations with a batch size of $5$ as suggested by \cite{do2024prompt}. At each iteration, the number of new prompts sampled is $5$.
\end{itemize}

\paragraph{Models' hyperparameters.} The models' hyperparameters are presented below.

\begin{itemize}
  \item \textbf{ChatGPT.} We use \emph{gpt-3.5-turbo-1106} for our experiments. We use a window size of $1500$ and Nucleus Sampling \citep{holtzman2019curious} as our decoding strategy with a $p$ value of $1$. We use the system role as ``You are a helpful assistant!''.
  \item \textbf{Mistral-7B-it-v0.2.} We use a window size of $1500$, and Sampling decoding strategy \citep{holtzman2019curious} ($do\_sampling=True$). We load the model from Huggingface Transformers library \citep{wolf-etal-2020-transformers} with the model id is ``mistralai/Mistral-7B-Instruct-v0.2''. We do not set any explicit system role.
\end{itemize}

\section{Prompts and Prompting Analysis} \label{appdx:prompts}

\subsection{GPT-4o-Judge's Prompt} \label{sec:prompt-for-gpt4-judge}

\llong{Our GPT-4o-Judge prompt evaluating the generated response and the reference is heavily motivated by \citet{zheng2023judging}.}

\begin{tcolorbox}[colback=white,coltext=black,boxrule=0.5pt,fonttitle=\small, fontupper=\small]
Please act as an impartial judge and evaluate how well an assistant's answer aligns with the reference answer and the quality of the assistant's answer. You will be given a user prompt, a reference answer and an assistant's answer. 
Your evaluation must consider the following criteria: 

- Format consistency: ensuring the generated response matches the length and structure of the reference.

- Content completeness: evaluating whether all key points present in the reference are included in the assistant's answer.

- Factuality: checking for factual correctness of the assistant's answer.

- Style adherence: ensuring that the tone, style, and level of detail of the of the assistant's answer match the reference.

- Assistant's answer quality: assessing how well the response satisfies the user's requirements.

Begin your evaluation by providing a short explanation for each. Be as objective as possible. After providing your explanation, please rate the response on all the criterion on a scale of 1 to 10 by strictly following this format: 

[The Start of Explanation]

...

[The End of Explanation]

[The Start of Ratings]

\{

"Format": 1-10,

"Content": 1-10,

"Factuality": 1-10,

"Style": 1-10,

"Quality": 1-10,

\}

[The End of Ratings]

[User Prompt]

{{user\_prompt}}

[The Start of Reference Answer]

{{answer\_ref}}

[The End of Reference Answer]

[The Start of Assistant’s Answer]

{{answer\_a}}

[The End of Assistant’s Answer]
\end{tcolorbox}

\subsection{ChatGPT Property Scorer Prompt} \label{sec:chatgpt-property-scorer}

\begin{tcolorbox}[colback=white,coltext=black,boxrule=0.5pt,fonttitle=\small, fontupper=\small]

You are an expert in evaluating the quality of a text generation task. You possess a nuanced understanding of various critical aspects. Brevity is paramount for you, ensuring concise expression without sacrificing essential information. Clarity is essential for comprehension, ensuring that your text is easily understood by the intended audience. Relevance ensures that the generated content aligns closely with the given context or prompt. Neutrality is crucial, maintaining an impartial tone devoid of bias. Coherence ties together ideas seamlessly, fostering a logical flow within your text. Completeness guarantees that all relevant points are addressed adequately. Specificity enhances precision, providing detailed and accurate information. Respect of chronology ensures temporal coherence, maintaining the chronological order of events. Accuracy demands factual correctness, avoiding errors or misinformation. Non-repetitiveness prevents redundancy, ensuring freshness in your expression. Indicative language aids in signaling key points or conclusions. Lastly, resolution ensures that your text concludes satisfactorily, resolving any questions or issues raised throughout.

Input: \{dialogue\}

Output: \{generated\_summary\}

Your task is to evaluate the following criteria in a scale of 1-5, with 1 is worst and 5 is best.

\{
    
    "Semantic Coverage": 1-5,
    
    "Factuality": 1-5,
    
    "Consistency": 1-5,
    
    "Informativeness": 1-5,
    
    "Coherence": 1-5,
    
    "Relevance": 1-5
    
\}

The definitions of the criteria are:

Semantic Coverage (COV): The extent to which a dialogue summary captures the main ideas and topics discussed in the conversation.

Factuality (FAC): The accuracy and truthfulness of the information presented in the dialogue summary, reflecting fidelity to the original conversation.

Consistency (CON): The degree to which the summary maintains logical and contextual coherence throughout, avoiding contradictory or conflicting information.

Informativeness (INF): The richness and depth of information conveyed in the dialogue summary, including key details and relevant context.

Coherence (COH): The overall clarity and organization of the summary, ensuring smooth transitions between ideas and coherence in the narrative flow.

Relevance (REL): The pertinence of the information included in the dialogue summary to the intended purpose or topic, ensuring alignment with the user's interests or needs.

Your output must be in Python dictionary format.

\end{tcolorbox}

\subsection{\model{}'s Prompts} \label{appx:longguide-prompts}

\paragraph{Prompting templates for \model{}.} 
Let $Q, C, I, D_f$ be the input query, context,  instruction, and demonstration token sequence respectively (\Cref{sec:intro}, \Cref{sec:demos-anayn}), and $G^{best}$ is the learned guideline(s), the prompt for $\mathcal{M}$ is formatted: \texttt{``\{$I$\}\textbackslash n\{$D_f$\}\textbackslash n\{$C$\}\textbackslash n\{$Q$\}\textbackslash n\{$G^{best}$\}''}. 

\paragraph{Prompting costs.}

\begin{table*}[hp!]
\centering
\footnotesize
\scalebox{.63}{
\begin{tabular}{cl|ccc|c|c|c|c}
\toprule
 & &  & \textbf{Summarization} & & \textbf{Simplification} & \textbf{Translation} & \textbf{Dialogue Generation}  &\textbf{Table2Text} \\
\midrule
\textbf{Models} & \textbf{Method} & \textbf{SAMSum} & \textbf{CNN (3.0.0)} & \textbf{XL-Sum} & \textbf{SWiPE} & \textbf{IWSLT17 en-ja} & \textbf{Synthetic Persona} & \textbf{CommGen-Chall.} \\
\midrule
& \textbf{\#shots (random)} &  3 & 3 & 5 & 3 & 5 &  5 & 5 \\
\midrule
\multirow{4}*{\begin{tabular}[c]{@{}l@{}}{\rotatebox{90}{Mistral}}
\end{tabular}} 
& & & & & & & & \\
& \#tokens consumed & 642 & 1110 & 811 & 1020 & 915 & 855 & 939 \\
& US\$ consumed & 0 & 0 & 0 & 0 & 0 & 0 & 0 \\
& & & & & & & & \\
\midrule
\multirow{2}*{\begin{tabular}[c]{@{}l@{}} {\rotatebox{90}{ChatGPT}} \\
\end{tabular}} 
& & & & & & & & \\
& \#tokens consumed & 1866 & 7683 & 4863 & 2380 & 1370 & 1344 & 1272 \\
& US\$ consumed & insignificant & insignificant & insignificant & insignificant & insignificant & insignificant & insignificant \\
& & & & & & & & \\
\bottomrule
\end{tabular}}
\caption{
\small{Total number of tokens consumed and US\$ consumed for models to learn the metric guideline (MG) and output constraint guideline (OCG). } 
}
\label{table:prompting-costs}
\end{table*}

\begin{table}[!ph]
\centering
\resizebox{1\linewidth}{!}{
\begin{tabular}{cl|l|l}
\toprule
& \textbf{Method} & \textbf{\#Prompts Sampled} & \textbf{Cost} \\ 
\midrule
\multirow{3}*{\begin{tabular}[c]{@{}l@{}} {\rotatebox{90}{ZS}} \\
\end{tabular}}
& adv-ICL         & (3 iterations) $\times$ (1 instruction) $\times$ (5 variants) & 15 $\times$ prompt validation cost \\ 
& APO             & (5 iterations) $\times$ (15 prompts sampled) $\times$ (1 instruction) & 75 $\times$ prompt validation cost \\ 
& \model{}       & 4 prompts (MG, OCG, MG-OCG, No guideline) & \textbf{4} $\times$ prompt validation cost \\ \midrule
\midrule
\multirow{3}*{\begin{tabular}[c]{@{}l@{}} {\rotatebox{90}{FS}} \\
\end{tabular}}
& adv-ICL         & (3 iterations) $\times$ (3 demonstrations + 1 instruction) $\times$ (5 variants) & 60 $\times$ prompt validation cost \\ 
& APO             & (5 iterations) $\times$ (15 prompts sampled) $\times$ (3 demonstrations + 1 instruction) & 300 $\times$ prompt validation cost \\ 
& \model{}       & 4 prompts (MG, OCG, MG-OCG, No guideline) & \textbf{4} $\times$ prompt validation cost \\ \bottomrule
\end{tabular}}
\caption{Prompting cost comparison between PO methods and \model{} based on \# new prompts sampled to test over the validation set.}
\label{tab:prompt_costs_vs_baselines}
\end{table}

\Cref{table:prompting-costs} presents the total number of tokens consumed for models to learn the metric guidelines and output constraint guideline (OCG) for both models with the hyperparameters of \model{} specified in \Cref{Appx:baselines-details}. We observe that the number of tokens needed to learn the guidelines is insignificant, demonstrating that LongGuide is a cost-effective solution and potentially beneficial for a wide range of applications.

\llong{\Cref{tab:prompt_costs_vs_baselines} presents the prompting cost comparision between \model{} and other PO algorithms. We compare the number of new prompts sampled by each algorithm for validation set verification, as these prompts are the primary cost bottleneck in PO algorithms. We observe that \model{} is approximately at least \textbf{3.75} times cheaper than adv-ICL in both settings and \textbf{18.75} times cheaper than APO. For SAMSum, the validation of one prompt using 50 samples involves approximately 22K tokens, which incurs a cost of 0.02 USD as of November 19, 2024.} 

\paragraph{Prompt for step 1, metric selection.}
Below is the prompt we use for step 1 selecting metrics for a given task.

\begin{tcolorbox}[colback=white,boxrule=0.5pt,fonttitle=\small,fontupper=\small]
{Select top-5 metrics that are the most important from the list below to evaluate a special way of \{TASK\_NAME\}.
\{str(PRE\_DEFINED\_ASSESSMEN\_METRICS)\}.}

{Here are some demonstrations of the task \{TASK\_NAME\}:
\{DEMONSTRATION\_STRING\}.}

{Output your list of metrics in Python list format without any explanation: [...].}
\end{tcolorbox}

\paragraph{Prompt for step 2, metric score collection.}
Below is the prompt we use for step 2 for evaluating selected metrics on the task.

\begin{tcolorbox}[colback=white,boxrule=0.5pt,fonttitle=\small, fontupper=\small]

{You are given an input and an output of a \{TASK\_NAME\} task.}

{Input: \{input\}}

{Output: \{output\}}

{Your task is to evaluate the following criteria on a scale of 1-5, with 1 being worst and 5 being best.}

{\{EVALUATION\_FORMAT\}}

{The definitions of the criteria are:
\{METRICS\_DEFINITIONS\}}

{Your output must be in Python dictionary format without explanation.}
\end{tcolorbox}

\paragraph{Prompt for step 2, collecting metrics' definitions.}
Below is the prompt we use for step 2 collecting METRICS\_DEFINITIONS for step 2.

\begin{tcolorbox}[colback=white,boxrule=0.5pt,fonttitle=\small, fontupper=\small]
{Define the list of following metrics in details as the quality of the output expected for the \{TASK\_NAME\} task.}

{\{metrics\}}

{Give me the list in bullet points.}
\end{tcolorbox}

\paragraph{Prompt for step 3, generating metric guideline (MG).} Below is the prompt we use for step 3, generating the metric guideline (MG).

\begin{tcolorbox}[colback=white,boxrule=0.5pt,fonttitle=\small, fontupper=\small]
{Now you are given the following metrics: \{metrics\_string\} for the \{TASK\_NAME\} task.}

{Based on these scores on a scale of 5 for the quality of the output: \{str(metrics\_collected\_scores)\}, define the expected quality of the output for each metric in natural language. Give me the list in bullet points.}
\end{tcolorbox}

\section{Examples}

\begin{figure}
\centering
\includegraphics[width=.95\linewidth, trim={0cm 0cm 0cm 0cm},clip]{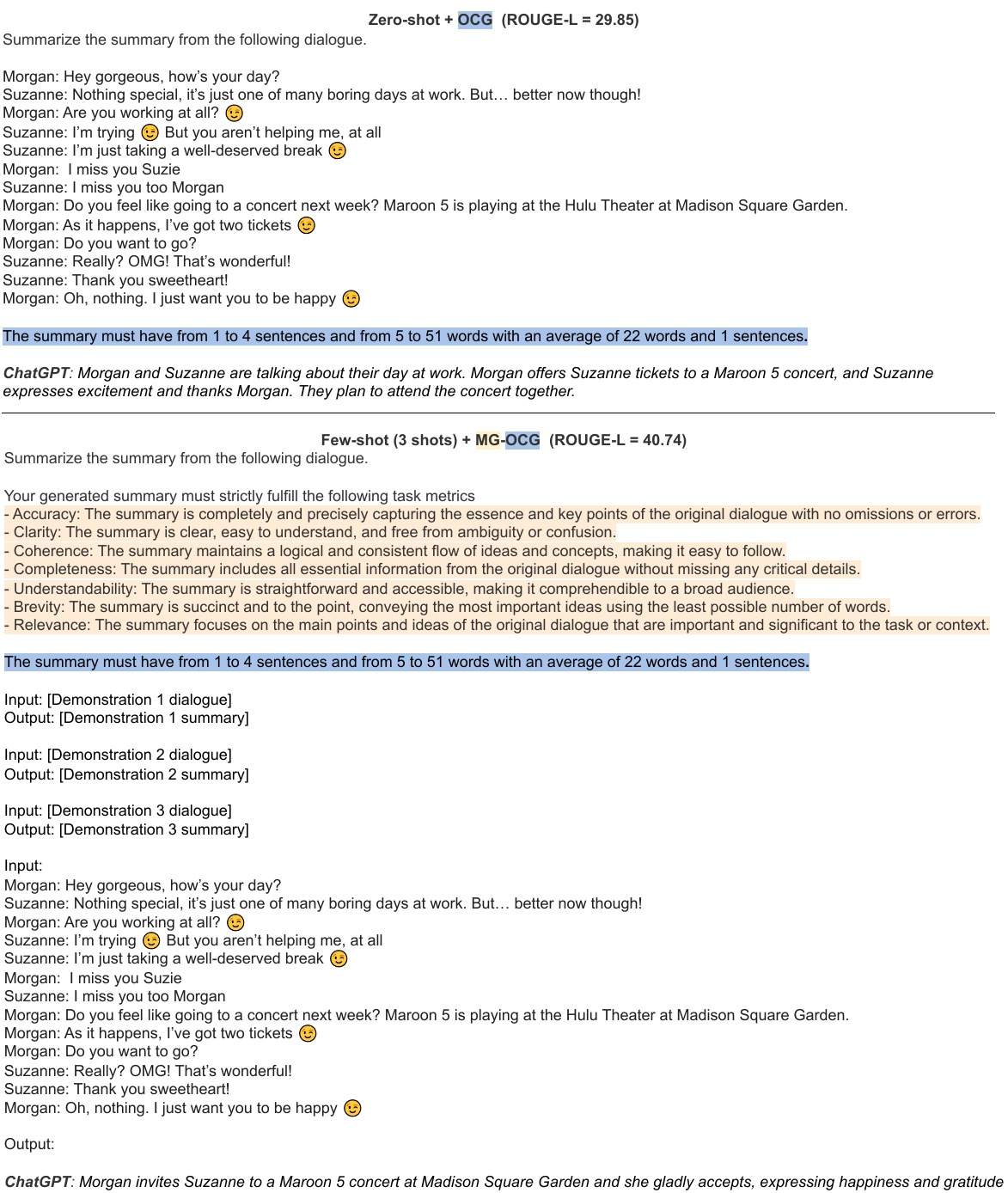}
\caption{\small{Full example of ChatGPT results on a SAMSum example \citep{gliwa-etal-2019-samsum}
w/ \model{} guidelines (Zero-shot + LongGuide and Few-shot + LongGuide) from \Cref{fig:longguide-teaser}}}\label{fig:teaser-full}
\end{figure}

\begin{figure}
\centering
\includegraphics[width=.8\linewidth, trim={0cm 0cm 0cm 0cm},clip]{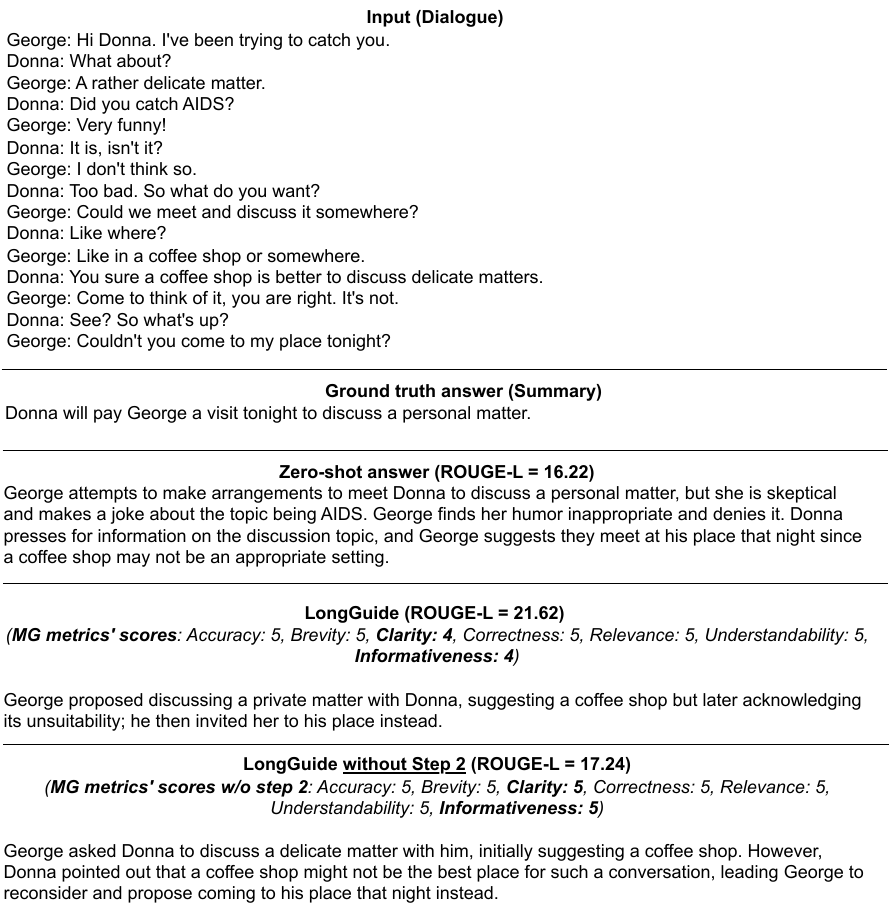}
\caption{\small{A SAMSum example, where skipping step 2 worsens the performance due to lack of clarity in metrics}}\label{fig:skip-step-2}
\vspace{-5mm}
\end{figure}

\begin{figure}
\centering
\includegraphics[width=\linewidth, trim={0cm 0cm 0cm 0cm},clip]{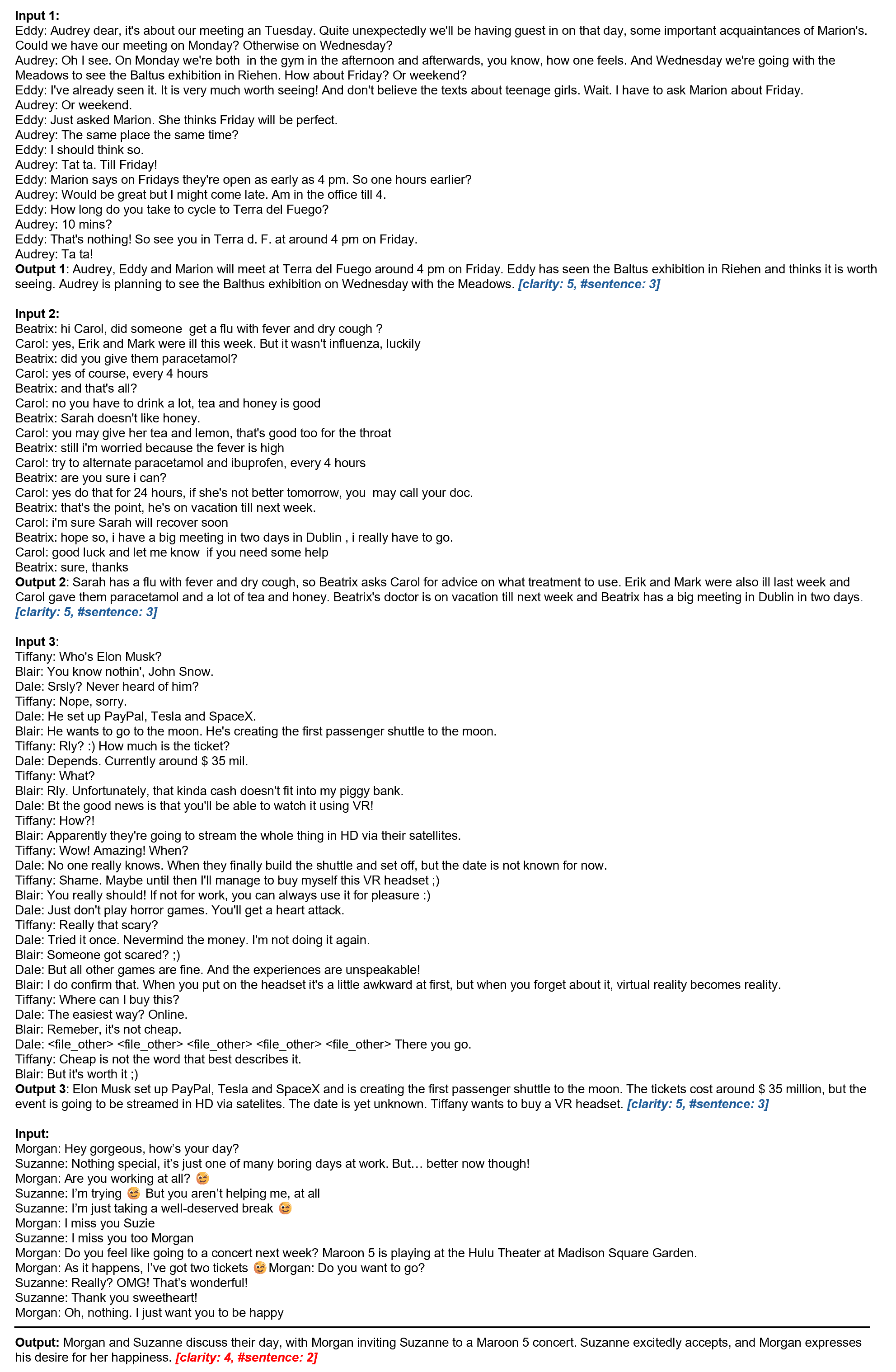}
\caption{\small{Full text for an example in \Cref{sec:demos-anayn}.}}\label{fig:full-example-theorem2.1}
\vspace{-5mm}
\end{figure}

\end{document}